\documentclass[runningheads]{llncs}

\usepackage{eccv}

\usepackage{eccvabbrv}

\usepackage{graphicx}
\usepackage{booktabs}

\usepackage[accsupp]{axessibility}  % Improves PDF readability for those with disabilities.

\usepackage{multirow}
\usepackage{colortbl}
\usepackage{mathtools}

\usepackage[pagebackref,breaklinks,colorlinks,citecolor=eccvblue]{hyperref}
\usepackage{orcidlink}

\definecolor{teal}{rgb}{0.0, 0.5, 0.5}
\definecolor{goldenrod}{rgb}{0.867, 0.769, 0.255}
\definecolor{gray}{rgb}{0.843, 0.843, 0.843}
\definecolor{brown}{rgb}{0.494, 0.259, 0.0196}
\definecolor{grey}{rgb}{0.9,0.9,0.9}

\newcommand{\methodname}{UniCal\xspace}
\newcommand{\truckdata}{\textit{MS-Cal}\xspace}
\newcommand{\pandaset}{\textit{PandaSet}\xspace}

\begin{document}

\title{UniCal: Unified Neural Sensor Calibration} 

\author{
Ze Yang\inst{1,2}\thanks{Indicates equal contribution. $^\dag$ Work done while an intern at Waabi.} \and
George Chen\inst{1,3}$^{\star\dag}$ \and
Haowei Zhang\inst{1} \and
Kevin Ta\inst{1} \and
Ioan Andrei Bârsan\inst{1,2} \and
Daniel Murphy\inst{1} \and
Sivabalan Manivasagam\inst{1,2} \and
Raquel Urtasun\inst{1,2}
}

\authorrunning{Z. Yang et al.}

\institute{
\mbox{
Waabi \quad \and
University of Toronto \quad \and
University of Waterloo
} \\
\email{\{zyang,gchen,hzhang,kta,abarsan,dmurphy,siva,urtasun\}@waabi.ai}
}

\maketitle

\begin{abstract}
Self-driving vehicles (SDVs) require accurate calibration of LiDARs and cameras to fuse sensor data accurately for autonomy.
Traditional calibration methods typically leverage fiducials captured in a controlled and structured scene and compute correspondences to optimize over.
These approaches are costly and require substantial infrastructure and operations, making it challenging to scale for vehicle fleets.
In this work, we propose \methodname, a unified framework for effortlessly calibrating SDVs equipped with multiple LiDARs and cameras.
Our approach is built upon a differentiable scene representation capable of rendering multi-view geometrically and photometrically consistent sensor observations.
We jointly learn the sensor calibration and the underlying scene representation through differentiable volume rendering, utilizing outdoor sensor data without the need for specific calibration fiducials.
This ``drive-and-calibrate'' approach significantly reduces costs and operational overhead compared to existing calibration systems, enabling efficient calibration for large SDV fleets at scale.
To ensure geometric consistency across observations from different sensors, we introduce a novel surface alignment loss that combines feature-based registration with neural rendering.
Comprehensive evaluations on multiple datasets demonstrate that \methodname outperforms or matches the accuracy of existing calibration approaches while being more efficient, demonstrating the value of \methodname for scalable calibration.
For more information, visit
    \href[]{https://waabi.ai/unical}{waabi.ai/unical}.
\end{abstract}
\section{Introduction}
\label{sec:intro}

\begin{figure*}[t]
    \centering
     \includegraphics[width=1.0\linewidth]{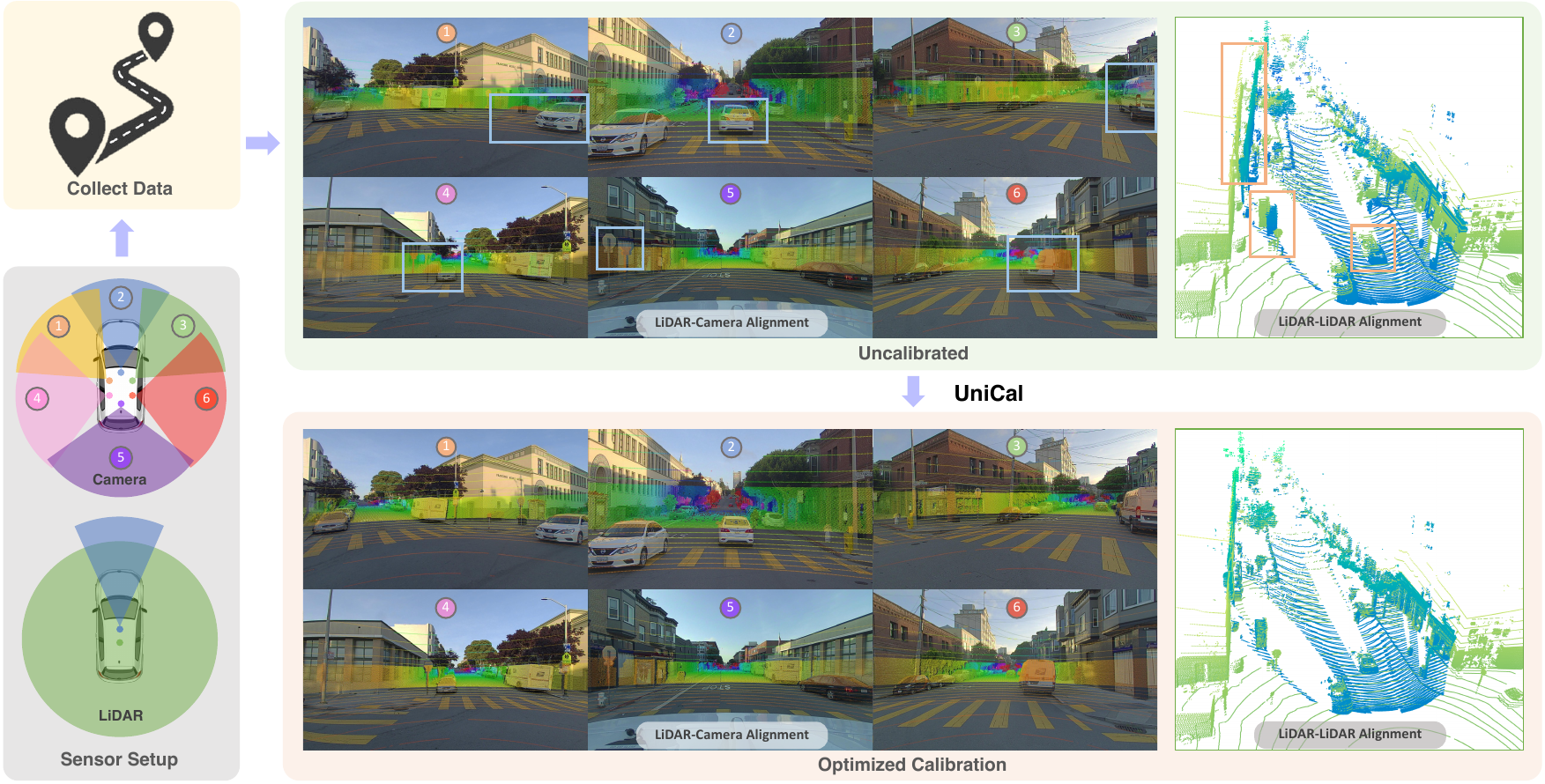}
     \caption{
     Our method takes collected data and automatically calibrates the sensor extrinsics.
     \textbf{Top}: LiDAR-Camera and LiDAR-LiDAR alignment on collected data with uncalibrated extrinsics.
     \textbf{Bottom}: Sensor alignment with our optimized calibration.
     }
     \label{fig:teaser}
\end{figure*}

Modern robotic systems, such as self-driving vehicles (SDVs), must observe the world accurately to perceive and plan safe actions.
To observe the world, they are equipped with a suite of sensors, including LiDARs and cameras, that provide depth and appearance information about their surroundings.
Modern SDVs require accurate sensor extrinsics to encode the relative poses between all sensors, to correctly interpret and process the observations in a shared coordinate frame, \eg, multi-sensor perception or 3D reconstruction.
Even a slight shift in the extrinsics estimation may result in several meters of misalignment between observations for distant objects, which could cause catastrophic failure.

Currently, multi-sensor extrinsics calibration in the self-driving industry is an arduous process that requires large infrastructure, significant operation costs, and substantial manual effort.
Typically, it involves collecting sensor data in a controlled indoor environment, with fiducials such as checkerboards mounted to fixed locations or held by operators \cite{fang2021single}.
The SDV must observe these fiducials from various viewpoints and distances. To enable this, turntables are often employed to ensure proper coverage (particularly in confined spaces) and repeatability \cite{jain2019high}.
This infrastructure is very expensive, and also incurs significant operation maintenance.
Traditional calibration methods~\cite{zhang2000flexible,zhang2004extrinsic,unnikrishnan2005fast} then leverage
extracted geometric features and exact checkerboard dimensions to compute correspondences and estimate the relative poses between different sensors.
Stage-wise calibration is often used to break down the problem: LiDAR-LiDAR, LiDAR-camera, and camera-camera pairs are first obtained,
followed by a global optimization stage such as pose graph optimization (PGO)~\cite{choi2015robust} or bundle adjustment (BA)~\cite{triggs2000bundle}.
Additionally, as sensors may shift after long periods of driving, SDVs have to be driven back to the facility for re-calibration every so often during normal operations~\cite{jain2019high}.
This complex process has multiple sources of error due to hardware, operations, and software that could result in poor calibration, such as warped fiducials~\cite{hagemann2022modeling}, insufficient observation of fiducials leading to ambiguity, or poor convergence of the calibration due to conflicting pose estimates during global optimization.
Such a calibration process has high cost and time overhead that makes it challenging to scale efficiently, \eg to multiple cities.

An ideal solution would instead rely on simply driving the SDV outdoors for a short period, running an algorithm, and automatically calibrating the entire multi-sensor extrinsics setup.
This ``drive-and-calibrate'' approach would dramatically reduce the cost and operations overhead compared to existing calibration systems and could calibrate large SDV fleets efficiently.
However, outdoor driving in an unstructured scene does not have clear fiducials that are often leveraged in classical calibration methods.
While there has been development of targetless calibration systems based on alignment of correspondence points or edges across sensor observations, they typically focus on a single sensor pair (e.g., LiDAR-camera~\cite{li2023automatic, pandey2012automatic}) or sensor modality~\cite{kim2020motion}, and may not achieve high quality calibration due to noisy matches.

Towards this goal, we propose \textbf{\methodname}, an automatic, targetless, multi-sensor calibration method based on neural rendering that computes extrinsics for an SDV equipped with multiple LiDARs and cameras.
We leverage the idea that replicating the sensor observations via neural rendering can provide a strong, dense supervision signal for calibration, without requiring targets or structured scene data.
\methodname enhances neural rendering specifically for multi-sensor calibration by incorporating a novel surface-guided alignment loss, and a coarse-to-fine sampling strategy based on robust feature correspondences, leading to improvements in both calibration and 3D reconstruction quality.

The main contributions of this paper are as follows:
(1) We propose a novel technique to calibrate LiDARs and cameras using neural fields without the need for specific calibration fiducials.
(2) We develop a surface alignment loss that combines feature-based registration with neural fields, as well as a progressive ray sampling schedule to improve the stability and quality of the optimization.
(3) We demonstrate the robustness and efficiency of our approach on two real-world autonomous driving datasets with both minivan and Class 8 truck sensor platforms and show that extrinsics obtained with \methodname lead to superior results.
Our findings demonstrate \methodname's value for scalable calibration.

\section{Related Work}
\label{sec:related}

\newcommand{\SE}[1]{\operatorname{\mathbb{SE}}(#1)}

\paragraph{Target-Based Multi-Sensor Calibration:}
The field of multi-sensor calibration has a decades-long history, with the majority of classic approaches relying on specially designed fiducials such as checkerboard patterns~\cite{zhang2004extrinsic,unnikrishnan2005fast,geiger2012automatic,zhou2018automatic}, custom 2D patterns ~\cite{alismail2012automatic,fang2021single,yan2023joint} or 3D patterns (cubes~\cite{chai2018novel}, spheres~\cite{ruan2014calibration,toth2020automatic}, holes~\cite{domhof2021joint}).
Identifying these patterns in both camera and LiDAR allows correspondences to be established, which can then be used to solve for the desired $\SE{3}$ transformation between the sensors.
Nevertheless, reliance on pre-existing fiducials hinders the large-scale deployment of robots and prevents the detection and correction of miscalibrations in the field, which may pose a safety hazard.
\methodname overcomes these challenges by not requiring calibration targets or a specific environment.

\paragraph{Targetless Multi-Sensor Calibration:}
Targetless methods use low-level signals, such as intensity and edge patterns in the environment to align cameras to LiDARs without explicit fiducials.
A recent survey~\cite{li2023automatic} classified this area into four sub-categories: (1) information-theory-based, (2) feature-based, (3) motion-based, and (4) learning-based. The first category maximizes an information-theoretic objective like mutual information to find the LiDAR-camera extrinsics~\cite{taylor2013automatic,pandey2015automatic,jiang2021semcal}.
Feature-based methods optimize the extrinsics by maximizing the alignment of LiDAR and camera features such as planes~\cite{li2023joint}, edges~\cite{levinson2013automatic,zhang2021line,yuan2021pixel, kang2020automatic}, or both~\cite{tu2022multi-camera}. Motion-based methods~\cite{ishikawa2018lidar} estimate 3D trajectories from camera and LiDAR data quasi-separately, and align the resulting trajectories to compute the transformation between the two sensors.
Finally, learning-based approaches~\cite{iyer2018calibnet,wu2021netcalib,lv2021lccnet,jing2022dxqnet,tarimofu2023batch} formulate extrinsic prediction from camera and LiDAR observations as a supervised learning task~\cite{schneider2017regnet}, possibly incorporating iterative refinement~\cite{iyer2018calibnet,lv2021lccnet} or differentiable optimization~\cite{tarimofu2023batch} in the network.
Hybrid methods have been proposed, for instance by first matching discrete features and then refining the estimate by maximizing a mutual information metric~\cite{koide2023general}.
Ou \etal~\cite{ou2023targetless} refine trajectory alignment results using LiDAR-camera feature matching.
These works have also been applied to online calibration \cite{aurora2023continuous, levinson2013automatic}.
By unifying neural rendering with calibration, \methodname implicitly leverages dense cross-sensor correspondences without the need to explicitly model low-level features.

\paragraph{Neural Rendering Pose Optimization:}
Neural Rendering has achieved rapid growth in several applications, such as view synthesis~\cite{mildenhall2021nerf,barron2021mip,barron2022mip,barron2023zip}, 3D reconstruction~\cite{wang2021neus,yariv2021volume,li2023neuralangelo,wang2023neural}, dynamic modelling~\cite{peng2021animatable,yang2021recovering,yang2021s3,pumarola2021d,cao2023hexplane}, and, recently, it has shown promise for autonomous driving simulation~\cite{yang2023unisim,tonderski2024neurad,pun2023lightsim,yang2023reconstructing,wang2023cadsim}.
The reliance of neural rendering methods on precise poses and calibration~\cite{mildenhall2021nerf} has led to a large body of work aiming to address this limitation, typically by extending the learning process to include refining the poses of the input images~\cite{wang2021nerf,lin2021barf,attal2021torf,meng2021gnerf,boss2022samurai,bian2023nope,smith2023flowcam,levy2023melon,heo2023robust}.
Jeong \etal~\cite{jeong2021self} further optimize camera intrinsics in addition to their poses.
While primarily studied through the lens of camera-only~\cite{wang2021nerf,lin2021barf,meng2021gnerf,boss2022samurai,smith2023flowcam,levy2023melon,heo2023robust} or
ToF-only~\cite{attal2021torf} reconstruction, calibration through neural fields has also been briefly explored for multimodal approaches~\cite{herau2023moisst,zhou2023inf}.
Recently, MOISST~\cite{herau2023moisst} proposed a neural rendering and calibration method which can account for extrinsic transformations between multiple cameras and LiDAR, as well as sensor time offsets.
SOAC~\cite{herau2023soac} improves convergence and robustness by learning per-camera fields and aligning them.
In contrast, \methodname leverages two novel calibration-focused elements, the surface alignment constraint and correspondence guided ray sampling, to improve estimation quality. 
Furthermore, we perform a rigorous experimental analysis in a multi-camera, multi-LiDAR setting.

\section{Unified Multi-Sensor Neural Calibration}
\label{sec:method}

% Math helpers
\newcommand{\bo}{\mathbf{o}}
\newcommand{\bd}{\mathbf{d}}
\newcommand{\bp}{\mathbf{p}}
\newcommand{\br}{\mathbf{r}}
\newcommand{\bx}{\mathbf{x}}
\newcommand{\bu}{\mathbf{u}}

\begin{figure*}[t]
    \centering
     \includegraphics[width=1.0\linewidth]{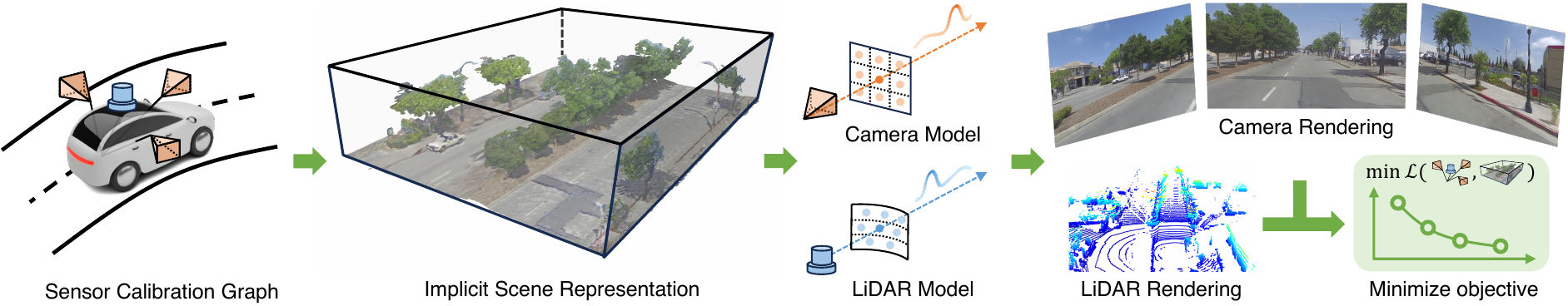}
     \caption{
     \textbf{Overview of our method.}
     We jointly optimize the multi-sensor extrinsics and underlying scene representation within a differentiable framework to minimize the photometric and geometric consistency losses on collected outdoor data retrospectively.
     }
     \label{fig:overview}
\end{figure*}

We propose a unified framework to calibrate the extrinsics of a multi-modal sensor platform which only requires collecting a short trajectory without the need for calibration targets.
We build our approach on a scene representation capable of rendering multi-view sensor observations in a geometrically and photometrically consistent manner.
Through the joint optimization of sensor parameters and the underlying scene representation within a differentiable framework, we effectively resolve the relationships between sensors.
Towards this goal, we introduce a novel differentiable surface alignment loss, which ensures geometric consistency across observations and mitigates the \textit{shape-radiance} ambiguity.
Our approach optimizes sensor extrinsics \wrt a vehicle reference (IMU) and assumes the intrinsics (e.g., focal length) as well as the vehicle trajectory are provided.
Note that this is a common setting in self-driving vehicles where localization is employed to estimate the latter.
In the remainder of this section, we first describe the neural scene representation in Sec.~\ref{sec:scene_repr}.
Then we detail the sensor rendering model in Sec.~\ref{sec:sensor_model}.
We present how to incorporate surface alignment constraints in Sec.~\ref{sec:surface_alignment}.
Finally we elucidate how to jointly estimate the calibration and scene representation in Sec.~\ref{sec:learning}.
Please see Fig.~\ref{fig:overview} for an overview of our method.

\subsection{Implicit Neural Scene Representation}
\label{sec:scene_repr}

Representing scene geometry and appearance using implicit representations~\cite{mildenhall2021nerf,mueller2022instant,mescheder2019occupancy} has gained considerable attention, due to their photorealistic results for novel view synthesis (NVS).
In this work, we leverage this scene representation to calibrate multiple cameras and LiDARs mounted on an SDV.
We parameterize the implicit representation using a multi-resolution feature grid with MLP networks~\cite{mueller2022instant}.
Specifically, given a 3D scene point $\mathbf x \in \mathbb R^3$, the 3D feature grid $\{ \mathbf G^l \}_{l=1}^L$ at each level is first tri-linearly interpolated.
The resulting interpolated features are then concatenated and processed with the MLP network to yield the geometry represented as a signed-distance function (SDF) $s$ and appearance feature $\mathbf f$.
This process can be characterized by a querying function $\mathcal Q$:
\begin{align}
\label{eqn:scene_repr}
	s, \mathbf f = \mathcal Q (\mathbf x) = \texttt{MLP}(\{ \texttt{interp}(\mathbf x, \mathbf G^l)_{l=1}^L \}).
\end{align}
In practice, we optimize the feature grid using a fixed number of features with a grid index hash function~\cite{mueller2022instant}.

\subsection{Differentiable Sensor Models}
\label{sec:sensor_model}
We focus on camera and LiDAR sensor calibration, as they are the primary sensory modalities employed by modern SDVs~\cite{sun2020scalability,wilson2021argoverse2}.
We denote the 6-DoF vehicle trajectory (IMU pose) expressed in an arbitrary world frame as $\mathbf P_\text{veh} (t)$.
Then, the state of each sensor at timestamp $t$ can be described as:
$\mathbf P_\text{sensor}^i (t) = \mathbf P_\text{veh} (t) ~ \mathbf E_\text{sensor}^i,$
where $\mathbf E_\text{sensor}^i$ represents the extrinsic matrix of the $i$-th sensor, indicating its relative pose \wrt the vehicle frame.

\paragraph{Camera Model:}
To render the camera image we represent each pixel $\mathbf u = (u, v)^T \in \mathbb R^2$ in homogeneous coordinates as $\mathbf{\bar u} = (u, v, 1)^T \in \mathbb R^3$.
Given the intrinsic matrix $\mathbf K_\text{cam}$ and extrinsic matrix $\mathbf E_\text{cam}$, we express the viewing ray in the 3D world coordinates as $\mathbf r(h_i) = \mathbf o + h_i \mathbf d$, where the ray origin $\mathbf o$ and ray direction $\mathbf d$ at timestamp $t$ are given by:
\begin{align}
\label{eqn:camera_ray}
	\mathbf o =
	\begin{bmatrix}
		\mathbf I & \mathbf 0
	\end{bmatrix}
	\mathbf P_\text{veh} (t) ~ \mathbf E_\text{cam}
	\begin{bmatrix}
		\mathbf 0 \\
		1
	\end{bmatrix},
	\quad
	\mathbf d =
	\begin{bmatrix}
		\mathbf I & \mathbf 0
	\end{bmatrix}
	\mathbf P_\text{veh} (t) ~ \mathbf E_\text{cam}
	\begin{bmatrix*}[l]
		\mathbf I \\
		\mathbf 0^\mathrm{T}
	\end{bmatrix*}
	\frac{\mathbf K_\text{cam}^{-1} ~ \mathbf{\bar u}}{\lVert \mathbf K_\text{cam}^{-1} ~ \mathbf{\bar u} \rVert_2}.
\end{align}
To render the image from the feature grid scene representation, the pixel color $\mathbf{\hat{I}}_\text{cam} (\mathbf r)$ can be approximated as the weighted sum of colors at sampled points $(1 \dots i \dots N)$ along the camera ray:
\begin{align}
\label{eqn:camera_render}
	\mathbf{\hat{I}}_\text{cam} (\mathbf r) &= \sum_{i=1}^N w_i \mathcal D_\text{cam} ( \mathbf f_i, \mathbf d ),
	\quad
	w_i = \alpha_i \prod_{j=1}^{i-1} (1-\alpha_j),
\end{align}
where $\alpha_i \in [0, 1]$ represents opacity, which we derive from the SDF $s_i$ of the $i$-th point following~\cite{yang2023unisim}.
The feature descriptor $\mathbf f_i$ and SDF $s_i$ are obtained from the querying function $\mathcal Q(\mathbf o + h_i \mathbf d)$ (Eq.~\eqref{eqn:scene_repr}), where $h_i$ is the $i$-th sample point depth along the ray, and
$\mathcal D_\text{cam}(\cdot)$ serves as the camera decoder, mapping the feature descriptor and view direction to RGB color.

\paragraph{LiDAR Model:}
The LiDAR sensor emits laser beam pulses and determines the distance from the sensor to the reflective surface by measuring the time of flight.
With the laser elevation and azimuth angles denoted as $\mathbf w = (\theta, \gamma)$ for each emitted ray, the ray direction in sensor coordinates can be characterized as $(\cos \theta \cos \gamma, \cos \theta \sin \gamma, \sin \theta)^\mathrm{T}$.
To model ego-motion during the LiDAR scan, we interpolate the vehicle pose $\mathbf P_\text{veh} (t)$ for each laser ray at firing timestamp $t$ and obtain the viewing ray origin and direction in the 3D world coordinates as:
\begin{align}
\label{eqn:lidar_ray}
	\mathbf o =
	\begin{bmatrix}
		\mathbf I & \mathbf 0
	\end{bmatrix}
	\mathbf P_\text{veh} (t) ~ \mathbf E_\text{lidar}
	\begin{bmatrix}
		\mathbf 0 \\
		1
	\end{bmatrix},
	\quad
	\mathbf d =
	\begin{bmatrix}
		\mathbf I & \mathbf 0
	\end{bmatrix}
	\mathbf P_\text{veh} (t) ~ \mathbf E_\text{lidar}
	\begin{bmatrix*}[l]
		\mathbf I \\
		\mathbf 0^\mathrm{T}
	\end{bmatrix*}
	\begin{bmatrix}
	\cos \theta  \cos \gamma \\
	\cos \theta  \sin \gamma \\
	\sin \theta
	\end{bmatrix}.
\end{align}
Similar to Eq.~\eqref{eqn:camera_render}, we apply differentiable volume rendering to generate the depth and intensity~\cite{yang2023unisim,Huang2023nfl}:
\begin{align}
\label{eqn:lidar_render}
	\hat{D} (\mathbf r) &= \sum_{i=1}^N w_i h_i,
	\quad
	\mathbf{\hat{I}}_\text{lidar} (\mathbf r) = \sum_{i=1}^N w_i \mathcal D_\text{lidar} ( \mathbf f_i, \mathbf d),
\end{align}
where $h_i$ represents the depth of the $i$-th sampled point, $\mathcal D_\text{lidar}(\cdot)$ denotes the LiDAR decoder, which maps the queried feature descriptor and view direction to the LiDAR intensity value.

\subsection{Surface Alignment Constraint}
\label{sec:surface_alignment}

\begin{figure}[t]
    \centering
     \includegraphics[width=0.6\linewidth]{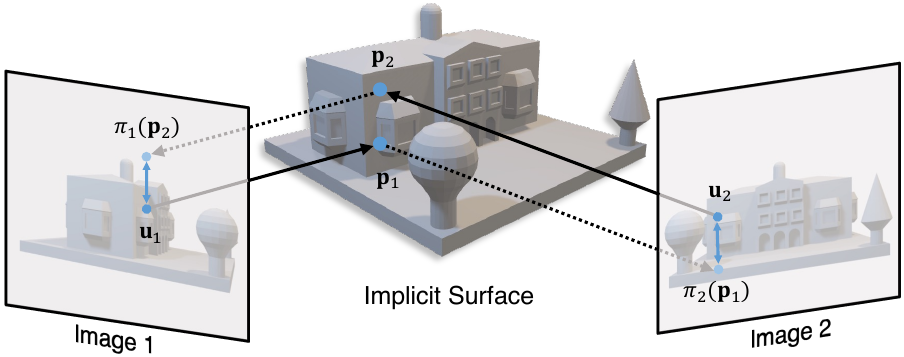}
     \caption{
     \textbf{Overview of surface alignment distance.}
     Ray-casting corresponding pixels $\mathbf{u}_1$ and $\mathbf{u}_2$ into the implicit surface yields 3D points $\mathbf{p}_1$ and $\mathbf{p}_2$.
     The surface alignment distance quantifies the image-space discrepancy between $\mathbf{p}_1$ and $\mathbf{p}_2$, and minimizing it ensures geometric consistency across sensors or perspectives.
     }
     \label{fig:surface_alignment_loss}
\end{figure}

Recovering both the scene representation and sensor poses can be challenging for unstructured outdoor driving scenes.
An unregularized model can learn to render the target observations with incorrect poses and geometry.
To mitigate this, we want to ensure that the 3D structures inferred from the sensor data aligns with the underlying implicit scene surface.
For LiDAR data, this geometric alignment can be assessed by comparing the rendered depth with the observed depth.
However, establishing geometric alignment for camera data is non-trivial due to the absence of direct depth information.

In this work, we introduce a differentiable surface alignment distance that provides additional geometric constraints on the camera poses.
Towards this goal, we infer sparse correspondences between pairs of camera images using off-the-shelf multi-view geometry tools~\cite{lindenberger2023lightglue}, denoting a pair of corresponding pixels between image pair as $(\bu_1, \bu_2)$.
The associated camera rays are computed from Eq.~\eqref{eqn:camera_ray} as $\br_1 (h) = \mathbf o_1 + h \mathbf d_1$ and $\mathbf r_2 (h) = \mathbf o_2 + h \mathbf d_2$ respectively.
In the case of perfect calibration, the rays $\mathbf r_1$ and $\mathbf r_2$ should precisely intersect at the scene surface.
Calibration inaccuracies introduce errors that can be measured by computing the distance between the ray-casted points on the scene surface.
The continuous nature of our scene representation allows us to query depth values for arbitrary rays within the scene using the depth function $\hat{D}(\cdot)$ defined in Eq.~\eqref{eqn:lidar_render}; the ray-casted points can then be expressed as:
\begin{align}
\label{eqn:ray_cast_surface}
	\mathbf p_1 = \mathbf o_1 + \hat{D}(\mathbf r_1) \mathbf d_1, \quad \mathbf p_2 = \mathbf o_2 + \hat{D}(\mathbf r_2) \mathbf d_2.
\end{align}
We normalize the distance between the ray-casted points by projecting them onto the image plane, and define the surface alignment distance as:
\begin{align}
\label{eqn:surface_distance}
	\mathcal \ell_\text{surf} &= \lVert \pi_1(\mathbf p_2) - \mathbf u_1 \rVert_2 + \lVert \pi_2(\mathbf p_1) - \mathbf u_2 \rVert_2,
\end{align}
where $\pi$ is the projection operator defined by camera intrinsic and extrinsic parameters.
Please see Fig.~\ref{fig:surface_alignment_loss} for an illustration of the surface alignment distance.

\subsection{Learning Sensor Calibration}
\label{sec:learning}
We jointly optimize the scene representation $\mathcal Q$, the decoders $\mathcal D_\text{cam}(\cdot)$, $\mathcal D_\text{lidar}(\cdot)$ and the sensor extrinsics $\mathbf E_\text{sensor}^i$ on a set of camera images and LiDAR sweeps to minimize the photometric and geometric losses.
Additionally, we also regularize the underlying scene geometry to ensure smooth surfaces and to satisfy physical constraints.
Our full learning objective is:
\begin{align}
	\mathcal L = \lambda_\text{photom} \mathcal L_\text{photom} + \lambda_\text{geom} \mathcal L_\text{geom} + \lambda_\text{reg} \mathcal L_\text{reg}.
	\label{eqn:objective}
\end{align}
We now elaborate on each loss term in more detail and present how we sample the rays to facilitate learning.

\paragraph{Photometric Consistency Loss:}
The photometric consistency loss $\mathcal L_\text{photom} = \mathcal L_\text{rgb} + \lambda \mathcal L_\text{int}$ consists of the camera RGB rendering loss and LiDAR intensity rendering loss.
The camera rendering loss is an $\ell_2$ loss calculated between the observed camera images and rendered images:
\begin{align}
\label{eqn:rgb_photo_loss}
	\mathcal L_\text{rgb} = \sum_{i=1}^{N_\text{cam}} \sum_{j=1}^N \left\lVert \mathbf{\hat{I}}_\text{cam}(\mathbf u_j^i \vert \mathbf{E}_\text{cam}^i) - \mathbf{I}_\text{cam} (\mathbf u_j^i) \right\rVert_2,
\end{align}
where $N_\text{cam}$ is the number of camera sensors and $\mathbf E_\text{cam}^i$ is the $i$-th camera sensor extrinsic, $\mathbf u_j^i$ is the sampled pixel from the images captured by the $i$-th camera sensor.
$\mathbf{\hat{I}}_\text{cam}$ is the rendered image from Eq.~\eqref{eqn:camera_render} and $\mathbf{I}_\text{cam}$ is the corresponding observed camera image.
Similarly, the LiDAR intensity rendering loss is:
\begin{align}
\label{eqn:intensity_photo_loss}
	\mathcal L_\text{int} = \sum_{i=1}^{N_\text{lidar}} \sum_{j=1}^N \left\lVert \mathbf{\hat{I}}_\text{lidar}(\mathbf w_j^i \vert \mathbf{E}_\text{lidar}^i)\,-\,\mathbf{I}_\text{lidar} (\mathbf w_j^i) \right\rVert_2,
\end{align}
where $\mathbf E_\text{lidar}^i$ is the $i$-th LiDAR sensor extrinsic and $N_\text{lidar}$ is the number of LiDAR sensors, $\mathbf w_j^i$ is the sampled laser beam.
$\mathbf{\hat{I}}_\text{lidar}$ (Eq.~\eqref{eqn:lidar_render}) and $\mathbf{I}_\text{lidar}$ are the rendered and observed LiDAR intensities.

\paragraph{Geometric Consistency Loss:}
The geometric consistency loss $\mathcal L_\text{geom} = \mathcal L_\text{depth} + \beta \mathcal L_\text{align}$ consists of the depth rendering loss and the surface alignment loss.
The depth rendering loss is an $\ell_1$ term between the rendered LiDAR depth and the observed LiDAR depth:
\begin{align}
\label{eqn:depth_loss}
	\mathcal L_\text{depth} = \sum_{i=1}^{N_\text{lidar}} \sum_{j=1}^N \left\lVert \hat{D}(\mathbf w_j^i \vert \mathbf E_\text{lidar}^i) - D(\mathbf w_j^i) \right\rVert_1,
\end{align}
where $\hat{D}(\cdot)$ is the depth rendering function defined in Eq.~\eqref{eqn:lidar_render}.
For surface alignment loss, we leverage an off-the-shelf image matching~\cite{lindenberger2023lightglue} algorithm to robustly identify corresponding pixels between image pairs.
Denoting the set of correspondences between camera $i$ and camera $j$ as $\{\mathbf u_k^i\}_{k=1}^M$ and $\{\mathbf u_k^j\}_{k=1}^M$, where $M$ is the number of correspondences, the surface alignment loss is then defined as:
\begin{align}
\label{eqn:alignment_loss}
	\mathcal L_\text{align} =
	\sum_{i=1}^{N_\text{cam}} \sum_{j=1}^{N_\text{cam}}
	{\sum_{k=1}^{M} \frac{\ell_\text{surf} \left(\mathbf u_k^i, \mathbf u_k^j \vert \mathbf E_\text{cam}^i, \mathbf E_\text{cam}^j\right)}{M}},
\end{align}
where $\ell_\text{surf}(\cdot)$ is the surface alignment distance defined in Eq.~\eqref{eqn:surface_distance}.
As image matching and ray-casting results may contain noise, we filter out correspondences with a re-projection error $\lVert \pi(\mathbf p) - \mathbf u \rVert_2 > \delta_\pi$, or a ray termination probability $\sum_{i=1}^N w_i < \delta_w$ and focus on reliable correspondences.

\paragraph{Regularization Term:}
We impose additional constraints on the learned surface representation.
Firstly, we promote concentration of the learned sample weight distribution $w_i$ around the surface.
Secondly, we encourage the underlying SDF value to satisfy the Eikonal equation, which helps the network to learn a smooth zero level set \cite{sitzmann2020implicit}.
This results in the following two terms:
\begin{align}
\label{eqn:regularizer_loss}
	\mathcal L_\text{reg} = \sum_{\tau_i>\epsilon} \lVert w_i \rVert_2
	+ \lambda \sum_{\tau_i<\epsilon} \left( \lVert \nabla s(\mathbf x_i) \rVert_2 - 1 \right)^2,
\end{align}
where $\tau_i = \lvert h_i - D \rvert $ represents the distance between the sample point $\mathbf x_i$ and its corresponding LiDAR depth observation.
These two terms contribute to learning an accurate surface representation, which yields precise depth values for optimizing the surface alignment loss $\mathcal{L}_\text{align}$.

\paragraph{Ray Sampling Priors:}
Selecting which sensor rays to render and supervise with is an important design choice for learning sensor calibration.
Typical structure-from-motion pipelines~\cite{schoenberger2016colmap} identify interest points to establish correspondences for alignment.
In contrast, the neural rendering literature~\cite{mildenhall2021nerf} typically leverages uniform ray sampling for scene reconstruction.
Our approach takes the best of both by employing a coarse-to-fine sampling strategy during training.
Initially, we uniformly sample sensor rays to learn an accurate scene representation.
However, not all sensor rays contribute equally to pose learning: textureless regions, like the sky and road, offer insufficient gradients to effectively update the sensor poses.
Therefore, we progressively increase sampling frequency in regions of interest to enhance pose registration.
Specifically, we identify interest points using an off-the-shelf detector~\cite{detone2018superpoint} and create a corresponding heat map $h$.
Denoting $\beta \in [0, 1]$ as the controllable parameter proportional to the progress in the coarse-to-fine sampling stage, the sampling probability map $p(\beta)$ is then proportional to the Gaussian-blurred version of the heat map:
\begin{align}
\label{eqn:sampling}
	p (\beta) &\propto h_\text{min} + \beta \cdot \texttt{GaussBlur}(h, k_\beta),
\end{align}
where $h_\text{min}$ controls the minimum score for sampling any rays, $k_\beta = \beta k_\text{min} + (1-\beta) k_\text{max}$ is the Gaussian blur kernel.

\section{Experiments}
\label{sec:experiments}

In this section, we introduce our experimental setting to evaluate \methodname.
We then compare our model against state-of-the-art methods across different driving scenes.
We also perform a comprehensive ablation on our design choices that enhance \methodname for calibration.
In the supplementary, we show that our improved calibration enables more realistic reconstruction and simulation of driving scenes.
We also study the calibration performance under various initialization and driving patterns, alongside comparisons with learning-based methods~\cite{iyer2018calibnet,lv2021lccnet}.

\begin{figure}[t]
    \centering
     \includegraphics[width=1.0\linewidth]{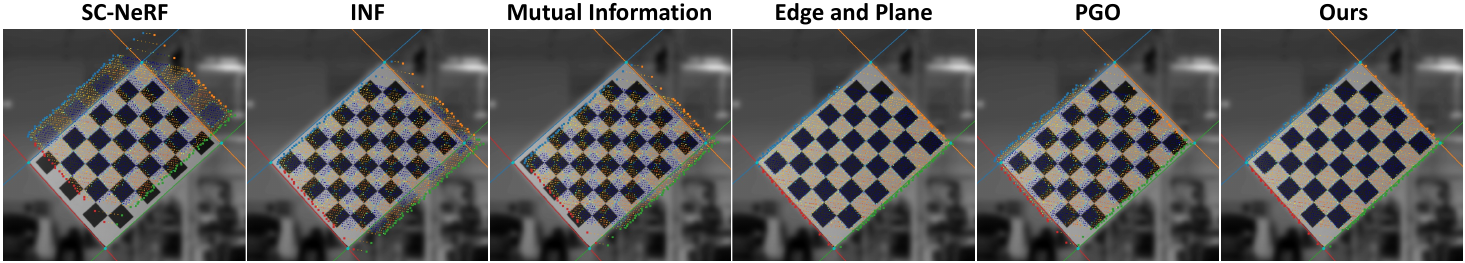}
     \caption{
     \textbf{Visualization of LiDAR-Camera alignment} on the checkerboard data.
     LiDAR points are colored with intensity value.
     }
     \label{fig:qual_checkerboard}
\end{figure}

\begin{figure}[t]
    \centering
     \includegraphics[width=1.0\linewidth]{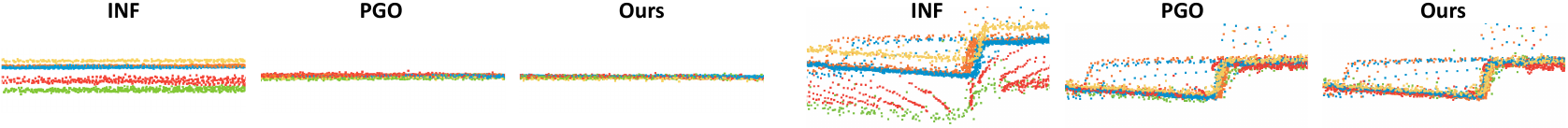}
     \caption{
     \textbf{Visualization of LiDAR-LiDAR alignment} on the flat ground and curb, with each LiDAR represented by a different color.
     }
     \label{fig:qual_lidar}
\end{figure}

\subsection{Experimental Setup}

\begin{figure*}[t]
    \centering
     \includegraphics[width=1.0\linewidth]{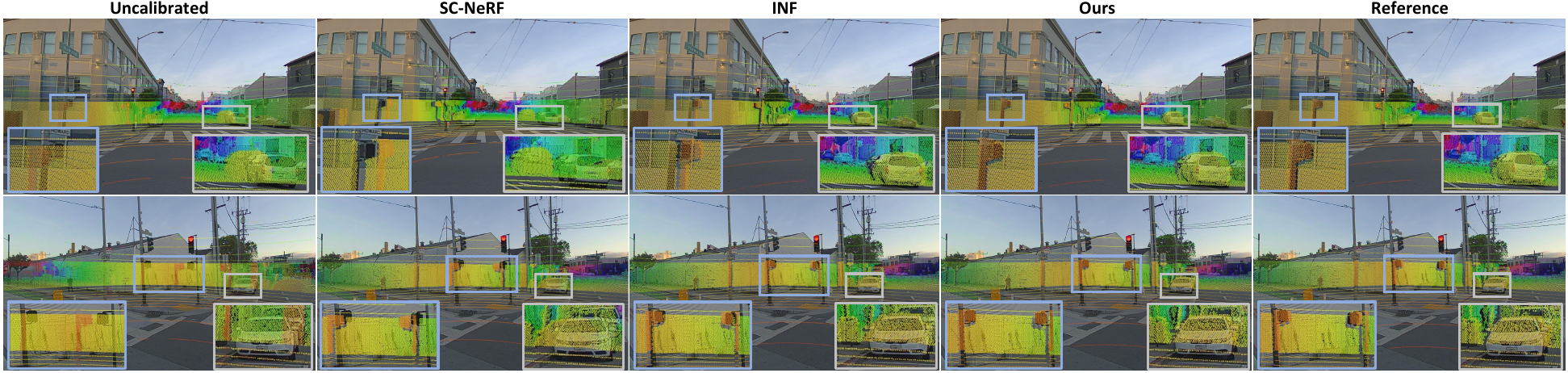}
     \caption{
     \textbf{Qualitative comparison on \pandaset.} 
     We show the projection of LiDAR and camera images and highlight mis-calibrations.
     }
     \label{fig:qual_pandaset}
\end{figure*}

\begin{figure*}[t]
    \centering
     \includegraphics[width=1.0\linewidth]{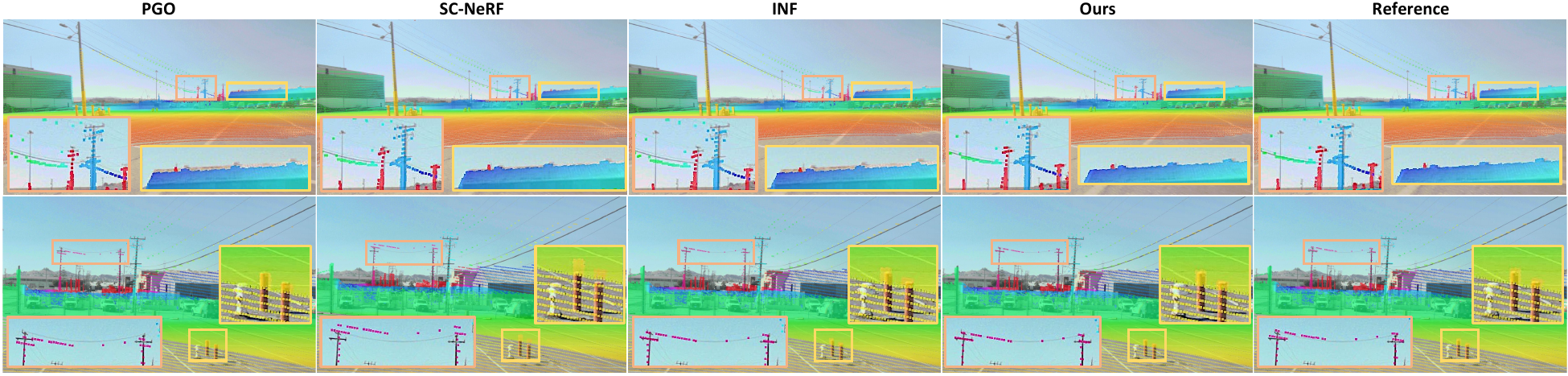}
     \caption{
     \textbf{Qualitative comparison on \truckdata dataset.} 
     We show the projection of LiDAR and cameras and highlight mis-calibrations.
     }
     \label{fig:qual_truck}
\end{figure*}

\paragraph{Multi-Sensor Calibration Dataset:}
To effectively evaluate \methodname against both classical target-based and target-free methods, and recent neural-rendering methods, we collected a multi-sensor \truckdata dataset consisting of stationary indoor data with checkerboard fiducials and several passes of outdoor driving data in a static parking lot.
The vehicle is a Class 8 truck equipped with five mechanical spinning LiDARs, (two long-range LiDARs (up to $200$m) and three medium-range LiDARs (up to $50$m)), and eight cameras (three wide-angle cameras, three medium-angle cameras, and two narrow-angle cameras arranged in a stereo pair).
For the indoor data, the SDV is stationary and the checkerboard is placed at various locations and orientations for camera-LiDAR pair alignment. 
One set of collects is used for optimizing the classical calibration methods, another set is held out for sensor alignment evaluation metrics.
For the outdoor parking lot data, we collect stationary and dynamic trajectories of the area, consisting of eight stationary scenes and four ``figure-8'' $\infty$ loops.
We select two $\infty$ loops to train neural-rendering methods and five stationary scenes for classical LiDAR alignment calibration.
We use the remaining $\infty$ loops and stationary scenes for evaluating the calibrations.
Please see supplementary for more details.

\paragraph{Urban Driving Dataset:}
We also evaluate our method on the publicly available real-world \pandaset~\cite{xiao2021pandaset}, which contains $103$ urban driving scenes captured in San Francisco, each lasting $8$ seconds.
The data collection platform consists of a $360^\circ$ mechanical spinning LiDAR and a forward-facing LiDAR, along with six cameras.
See Fig.~\ref{fig:teaser} for the sensor setup.
As we focus on calibration from static scenes, we select four scenes from \pandaset that have few dynamic actors to run the calibration, and select two scenes to evaluate the rendering performance.
We use 3D bbox labels to mask out the dynamic actor pixels and LiDAR points.

\paragraph{Baselines:}
We compare our method against both classical calibration approaches and NeRF-based pose estimation approaches.
For classical calibration approaches, we compare to \textbf{Point-to-Plane ICP}~\cite{chen1992object} and \textbf{MLCC}~\cite{liu2022targetless} for LiDAR-LiDAR calibration.
We also compare to LiDAR-camera calibration methods including aligning \textbf{Edge and Plane}~\cite{zhou2018automatic} correspondences on checkerboards and registering RGB and LiDAR intensity via \textbf{Mutual Information}~\cite{pandey2012automatic} on outdoor static collects.
Additionally, we compare to a multi-sensor calibration baseline based on global \textbf{Pose Graph Optimization} (PGO)~\cite{choi2015robust}.
Specifically, we construct a calibration graph for each LiDAR-LiDAR~\cite{chen1992object}, LiDAR-camera~\cite{zhou2018automatic}, and LiDAR-INS~\cite{furgale2010visual,barfoot2014associating} edge, and run PGO to align the full sensor setup.
For NeRF-based approaches, we compare to camera-only \textbf{SC-NeRF}~\cite{jeong2021self} and multi-sensor \textbf{INF}~\cite{zhou2023inf}.
We adapt the NeRF-based baselines to our setting by using the dataset-provided vehicle poses and exclusively optimizing the sensor to vehicle extrinsics.

\begin{table*}[t]
    \centering
    \resizebox{1.0\textwidth}{!}{
    \begin{tabular}{llcccccccccc}
    \toprule
    \multirow{2}{*}{Methods} & \multirow{2}{*}{Sensors} & \multicolumn{2}{c}{Multi-Sensor Alignment} & \multicolumn{2}{c}{Camera Pose Accuracy} & \multicolumn{2}{c}{LiDAR Pose Accuracy} & \multicolumn{4}{c}{Rendering Quality} \\
    \cmidrule(l){3-4} \cmidrule(l){5-6} \cmidrule(l){7-8} \cmidrule(l){9-12}
    & & LiD$\leftrightarrow$Cam$\downarrow$ & LiD$\leftrightarrow$LiD$\downarrow$ & Rotation$\downarrow$ & Transl$\downarrow$ & Rotation$\downarrow$ & Transl$\downarrow$ & PSNR$\uparrow$ & SSIM$\uparrow$ & LPIPS$\downarrow$ & Depth$\downarrow$ \\
    \midrule
    Point-to-Plane ICP~\cite{chen1992object} & LiD-LiD & - & \textbf{2.811} & - & - & 0.043 & 0.014 & - & - & - & -
    \\
    MLCC~\cite{liu2022targetless} & LiD-LiD & - & 3.064 & - & - & 0.068 & 0.022 & - & - & - & -
    \\
    Mutual Info~\cite{pandey2012automatic} & LiD-Cam & 39.76 & - & 1.182 & 0.200 & - & - & - & - & - & -
    \\
    Edge and Plane~\cite{zhou2018automatic} & LiD-Cam & 9.83 & - & 0.358 & 0.030 & - & - & - & - & - & -
    \\
    Pose Graph Optim~\cite{zhou2018open3d} & Full & 11.14 & 2.962 & 0.438 & \textbf{0.029} & 0.049 & 0.012 & 30.11 & 0.854 & 0.422 & 0.073
    \\
    \midrule
    SC-NeRF~\cite{jeong2021self} & Cam-Cam & 58.80 & - & 0.544 & 0.204 & - & - & 29.82 & 0.854 & 0.438 & -
    \\
    INF~\cite{zhou2023inf} & LiD-Cam & 47.27 & 8.996 & 0.645 & 0.189 & 0.368 & 0.116 & 30.77 & 0.872 & 0.400 & 0.148
    \\
    \rowcolor{grey} Ours & Full & \textbf{9.27} & 2.847 & \textbf{0.186} & 0.033 & \textbf{0.036} & \textbf{0.008} & \textbf{31.96} & \textbf{0.903} & \textbf{0.344} & \textbf{0.035}
    \\
    \bottomrule
    \end{tabular}
    }
	\caption{
	\textbf{State-of-the-art calibration comparison on \truckdata dataset.}
	Our method achieves best performance in terms of LiDAR-Camera re-projection error (pixel), LiDAR-LiDAR registration error (cm), pose rotation (degree), translation (m), and rendering quality.
	}
	\label{tab:sota_truck}
\end{table*}

\begin{table}[t]
    \centering
    \resizebox{0.65\textwidth}{!}{
    \begin{tabular}{lccccccc}
    \toprule
    \multirow{2}{*}{Methods} & \multicolumn{2}{c}{Camera Pose} & \multicolumn{2}{c}{LiDAR Pose} & \multicolumn{3}{c}{Rendering Quality}\\
    \cmidrule(l){2-3} \cmidrule(l){4-5} \cmidrule(l){6-8}
    & Rot$\downarrow$ & Tran$\downarrow$ & Rot$\downarrow$ & Tran$\downarrow$ & PSNR$\uparrow$ & SSIM$\uparrow$ & LPIPS$\downarrow$ \\
    \midrule
    SC-NeRF~\cite{jeong2021self} & 1.127 & 0.636 & - & - & 23.50	 & 0.672 & 0.504
    \\
    INF~\cite{zhou2023inf} & 0.767 & 0.350 & 0.136 & 0.126 & 23.64 & 0.689 & 0.489
    \\
    \rowcolor{grey} Ours & \textbf{0.267} & \textbf{0.122} & \textbf{0.048} & \textbf{0.015} & \textbf{25.14} & \textbf{0.727} & \textbf{0.450}
    \\
    \bottomrule
    \end{tabular}
    }
	\caption{
	\textbf{State-of-the-art calibration comparison on \pandaset.}
	Our method achieves best performance in terms of pose accuracy and rendering quality.
	}
	\vspace{-5pt}
	\label{tab:sota_panda}
\end{table}

\paragraph{Evaluation Metrics:}
We evaluate the performance of our proposed method in three different aspects: (1) classical multi-sensor alignment, (2) pose accuracy with respect to a reference, and (3) rendering quality at novel viewpoints.

\noindent (1) To measure \textbf{multi-sensor alignment}, we examine each LiDAR-camera pair and LiDAR-LiDAR pair that has overlap on the evaluation collect, and report the mean error.
For LiDAR-camera alignment, we follow~\cite{zhou2018automatic} and compute the reprojection error between the projected corners of the checkerboard planes from the LiDAR points and those in the images, measured in pixel space.
For LiDAR-LiDAR alignment, we compute the average Point-to-Plane distance (cm) of all inlier correspondences for each LiDAR pair on the stationary evaluation scenes.

\noindent (2) To assess \textbf{pose accuracy}, we report the average rotation and translation error between the reference and estimated calibrations.
For \pandaset~\cite{xiao2021pandaset}, their provided calibration serves as the reference.
For \truckdata dataset, we run blackbox optimization~\cite{powell1964efficient} to search for the calibration that minimizes both LiDAR-camera re-projection error and LiDAR-LiDAR registration error on the evaluation collect.
Please see supplementary for more details.

\noindent (3) For \textbf{rendering quality},
we train a fixed neural rendering algorithm \cite{yang2023unisim} using the generated calibration and report standard evaluation metrics~\cite{lin2021barf,jeong2021self}, including PSNR, SSIM~\cite{wang2004image} and LPIPS~\cite{zhang2018unreasonable} for camera rendering.
We report median ray depth error (m) for LiDAR rendering quality.

\paragraph{Calibration Initialization:}
The calibration methods optimize with respect to an initial estimate.
We evaluate two initialization settings: \emph{from-scratch}, where the initial calibration is a single reference point on the vehicle, and \emph{from-blueprint}, where the intial calibration is based on a CAD model blueprint used to install the sensors.
For \pandaset~\cite{xiao2021pandaset}, we evaluate with \emph{from-scratch} and initialize all the sensors at the location of the spinning LiDAR, and initialize the rotation of the 6 cameras with yaw angle of $0^\circ$, $45^\circ$, $90^\circ$, $180^\circ$, $270^\circ$, $315^\circ$, along with roll and pitch angles initialized to $0^\circ$.
The rotations for the 2 LiDARs are initialized with the identity matrix.
This setting is challenging, with a maximum rotation of $\approx 12.8^\circ$ and translation of $\approx 1.1$ meters against the reference.
For the \truckdata dataset, we initialize with the \emph{from-blueprint} setting.

\subsection{Calibration Evaluation}

\paragraph{Multi-Sensor Alignment:}
Our \truckdata dataset allows us to evaluate LiDAR-Camera pairwise re-projection error on the indoor checkerboard scenes as well as LiDAR-LiDAR pairwise registration error on the outdoor static scenes.
We show results in Table~\ref{tab:sota_truck}.
In comparison to classical calibration approaches, our method, which only used sensor data from the outdoor $\infty$-loop, achieves lower LiDAR-camera re-projection error on the checkerboard.
Regarding the LiDAR-LiDAR registration metric, ICP~\cite{chen1992object} serves as a strong baseline on the driving data, but after applying PGO to align multiple LiDARs in a global space, the performance drops, possibly due to conflicting pose edges during global optimization.
For NeRF-based baselines, since SC-NeRF~\cite{jeong2021self} does not calibrate LiDAR, we pre-align the optimized camera calibration to the reference and then use the reference LiDAR calibration to compute the LiDAR-Camera re-projection error.
Our method significantly outperforms the NeRF-based baselines and achieves better multi-sensor alignment.
Fig.~\ref{fig:qual_checkerboard} shows a visual comparison of the LiDAR-Camera alignment on checkerboard data and Fig.~\ref{fig:qual_lidar} shows a visual comparison of the LiDAR-LiDAR alignment on the ground.

\paragraph{Sensor Pose Accuracy:}
As several methods we compare against only calibrate sensors in a single modality and do not calibrate with respect to a reference point on the vehicle, we select a root sensor, align its calibrated pose with its reference pose, and then compute the average pose error.
Table~\ref{tab:sota_truck} and Table~\ref{tab:sota_panda} show the sensor pose accuracy on the \truckdata and the urban driving \pandaset~\cite{xiao2021pandaset}.
Our method achieves significantly lower camera and LiDAR rotation errors compared to all the baselines on both datasets, which is key for sensor fusion at range, demonstrating the effectiveness of our method.
Fig.~\ref{fig:qual_truck} and Fig.~\ref{fig:qual_pandaset} show the comparison of LiDAR projected to camera image with the estimated sensor pose on \truckdata dataset and \pandaset.
Our method achieves high sensor alignment even for thin and distant structures such as poles and powerlines.

\begin{figure}[t]
    \centering
     \includegraphics[width=1.0\linewidth]{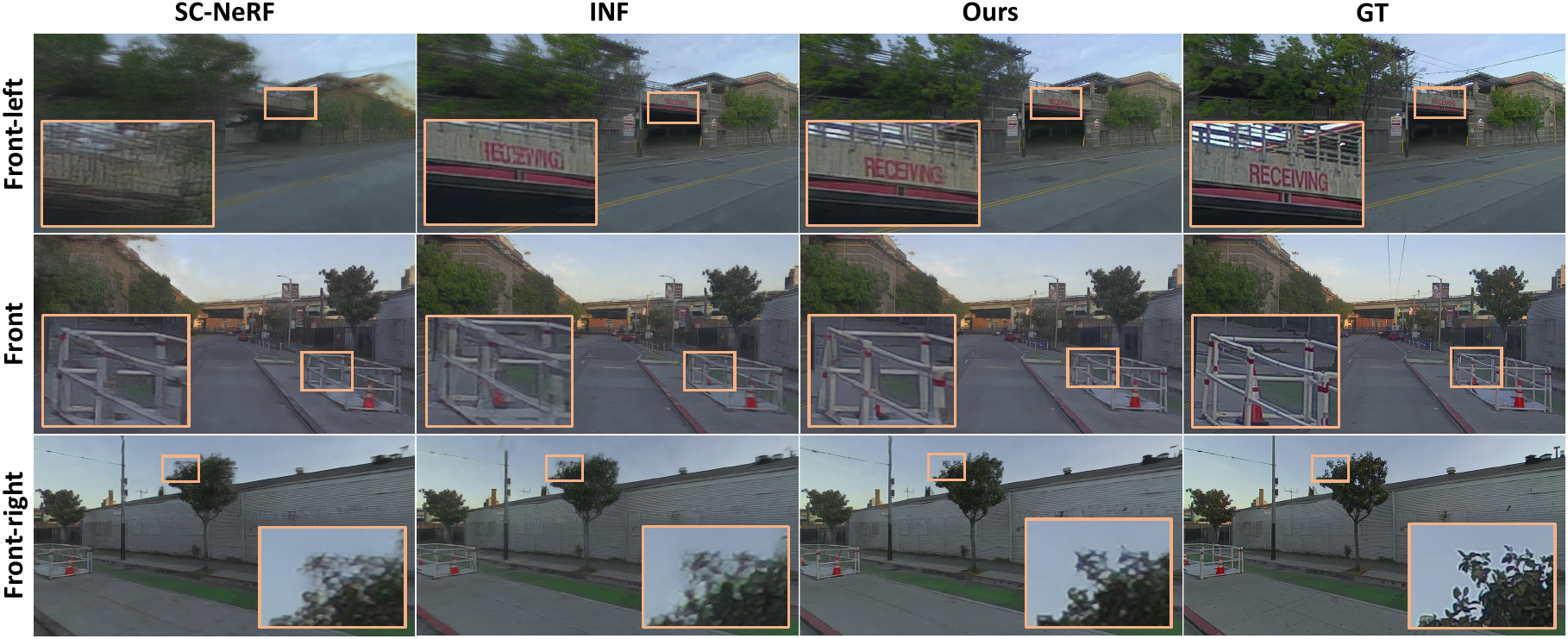}
     \caption{
     \textbf{Multi-camera rendering} using the calibration results.
     }
     \label{fig:qual_rendering}
\end{figure}

\begin{table}[t]
    \centering
    \resizebox{0.65\textwidth}{!}{
    \begin{tabular}{lccccccc}
    \toprule
    \multirow{2}{*}{Model} & \multicolumn{2}{c}{Camera Pose} & \multicolumn{2}{c}{LiDAR Pose} & \multicolumn{3}{c}{Rendering Quality}\\
    \cmidrule(l){2-3} \cmidrule(l){4-5} \cmidrule(l){6-8}
    & Rot$\downarrow$ & Tran$\downarrow$ & Rot$\downarrow$ & Tran$\downarrow$ & PSNR$\uparrow$ & SSIM$\uparrow$ & LPIPS$\downarrow$ \\
    \midrule
    Full & \textbf{0.267} & \textbf{0.122} & 0.048 & \textbf{0.015} & \textbf{25.14} & \textbf{0.727} & \textbf{0.450}
    \\
    w/o $\mathcal L_\text{align}$ & 0.856 & 0.614 & \textbf{0.047} & \textbf{0.015} & 23.89 & 0.681 & 0.500
    \\
    w/o \textit{M-S} & 0.469 & 0.216 & 0.049 & \textbf{0.015} & 24.38 & 0.698 & 0.477
    \\
    w/o $\mathcal L_\text{reg}$ & 0.288 & \textbf{0.122} & 0.050 & 0.016 & 25.08 & 0.725 & 0.453
    \\
    w/o \textit{I-S} & 0.278 & \textbf{0.122} & \textbf{0.047} & \textbf{0.015} & 25.08 & 0.725 & 0.454
    \\
    w/o $\mathcal L_\text{int}$ & 0.277 & 0.123 & 0.050 & \textbf{0.015} & 25.09 & 0.725 & 0.453 
    \\
    \bottomrule
    \end{tabular}
    }
	\caption{
	\textbf{Ablation of different components} on \pandaset.
	\textit{M-S} denotes training on \textit{M}ultiple \textit{S}ensors jointly, and \textit{I-S} denotes \textit{I}nterest region \textit{S}ampling.
	}
	\label{tab:ablation}
\end{table}

\paragraph{Rendering Quality:}
We also evaluate the scene reconstruction and novel view rendering with the generated calibration.
We run the calibration on the calibration collects, and then we train the scene reconstruction model~\cite{yang2023unisim} on the evaluation collects using the calibration result, and report the rendering metrics on the held-out frames.
Table~\ref{tab:sota_truck} and Table~\ref{tab:sota_panda} show the image rendering metrics and LiDAR depth rendering metric on our collected \truckdata and \pandaset~\cite{xiao2021pandaset}.
Our method beats the baselines in all rendering metrics.
Fig.~\ref{fig:qual_rendering} shows a visual comparison, where our method recovers text and thin structures more clearly.

\paragraph{Ablation Study:}
We study the effectiveness of several key components on \pandaset and present the results in Table~\ref{tab:ablation}.
The surface alignment loss $\mathcal L_\text{align}$ (Eqn.~\ref{eqn:alignment_loss}) is important to accurately recover the camera calibration on the challenging urban driving scenes.
We also compare to a model training each camera sensor separately, which is worse than of training on \textit{M}ultiple \textit{S}ensors jointly (\textit{M-S}), as it falls short in leveraging the multi-sensor constraints in overlap regions.
The regularizer term $\mathcal L_\text{reg}$ (Eqn.~\ref{eqn:regularizer_loss}) helps to learn accurate scene representation.
The \textit{I}nterest region \textit{S}ampling (\textit{I-S}) facilitates accurate sensor registration.
And the LiDAR intensity rendering loss $\mathcal L_\text{int}$ (Eqn.~\ref{eqn:intensity_photo_loss}) helps sensor matching.

\section{Conclusion}
\label{sec:conclusion}
In this paper, we propose \methodname, a unified framework that takes the collected data from multi-sensor platforms and automatically calibrates the sensor extrinsics.
Our method combines feature-based registration with neural rendering for accurate and efficient calibration without the need for calibration targets.
This ``drive-and-calibrate'' approach significantly reduces costs and operational overhead compared to existing calibration systems that employ extensive infrastructures and procedures, thereby facilitating scalable calibration for large SDV fleets.
Our method can also be combined with initial classical calibration approaches to further improve robustness.
Future work involves jointly learning the localization, sensor intrinsics, and modeling additional sensors.
Potential negative social impacts from calibration discrepancies can be mitigated through rigorous autonomy safety assessment prior to deployment.

\section*{Acknowledgements}
We thank Yun Chen, Jingkang Wang, Richard Slocum for their helpful discussions.
We also appreciate the invaluable assistance and support from the Waabi team.
Additionally, we thank the anonymous reviewers for their constructive comments and suggestions to improve this paper.

%\bibliographystyle{splncs04}
%\bibliography{main}

\newpage
\appendix
\renewcommand{\thesection}{A\arabic{section}}
\renewcommand{\thefigure}{A\arabic{figure}}
\renewcommand{\thetable}{A\arabic{table}}

\section*{Appendix}

In the supplementary material, we provide information about the implementation details of our method, along with details about the baselines, experimental settings, additional results and the discussion of limitations and potential negative social impact of our method.
We first describe the implementation details of our approach in \cref{sec:additional_details}.
Then in \cref{sec:baseline_details} we present additional details on the baseline model implementations and their adaptation to our setting.
Next, we explain in more detail our experimental setting and datasets in \cref{sec:exp_details}.
After that, we showcase additional experiments, including comaparison to additional learning-based (CalibNet~\cite{iyer2018calibnet}, LCCNet~\cite{lv2021lccnet}) and NeRF-based (MOISST~\cite{herau2023moisst}) calibration methods, additional results and visualizations in \cref{sec:additional_results}.
Finally, we analyze the limitations of our model and discuss the future works and potential negative social impact in \cref{sec:limitation}.

\section{\methodname Details}
\label{sec:additional_details}

\paragraph{Scene Representation Model:}
Our scene representation model is based on a multi-resolution feature grid and \texttt{MLP} network.
Following~\cite{mueller2022instant}, we employ a spatial hash function to map each feature grid to a fixed number of features, with the hash table size set to $2^{21}$.
To obtain the signed distance value $s$ and appearance feature $\mathbf f$ from the interpolated feature (\cref{eqn:scene_repr} in main paper), the \texttt{MLP} network consists of two layers, with a hidden size of $64$.
For distant regions outside the scene volume, we adopt an inverted sphere parameterization similar to NeRF++\cite{zhang2020nerf++}.

\paragraph{Camera and LiDAR Intensity Decoder:}
The camera RGB decoder $\mathcal D_\text{cam}$ (\cref{eqn:camera_render} in main paper) and the LiDAR intensity decoder $\mathcal D_\text{lidar}$ (\cref{eqn:lidar_render} in main paper) are both three layer MLPs.
They take the queried appearance feature $\mathbf f$ and view direction encoding $\mathbf d$ as input and output the RGB color and LiDAR intensity.
To account for variations in exposure and color tone across different sensors, we learn a per-sensor linear mapping over the intensity channel.
We found this simple model to be effective in capturing these variations in practice.

\paragraph{Rendering Details:}
To perform efficient volume rendering, we leverage the geometry priors from LiDAR observations to identify near-surface regions, enabling the evaluation of the radiance field exclusively within these areas.
This significantly reduces the number of required samples and radiance queries.
Specifically, we generate an occupancy grid for the scene volume using the aggregated LiDAR point clouds similar to~\cite{yang2023unisim}, with a voxel size set to be $0.5$ m.
We sample query points with a fixed step of $10$ cm for regions inside the scene volume, and sample an additional $16$ points for the distant sky region during volume rendering.

\paragraph{Sensor Pose Representation:}
The choice of sensor pose parameterization plays a crucial role for pose-optimizing NeRFs.
For instance, certain parameterizations of rotation, such as Euler angles, are known to lack continuity over the $\mathbb{SO}(3)$ manifold, posing challenges in the learning process.
To address this, we use a continuous 6D representation~\cite{zhou2019continuity} to parameterize the sensor rotation and 3D vector to parameterize the sensor translation.

\paragraph{Rolling Shutter Modelling:}
LiDAR sensors usually accumulate measurements over time, it takes non-negligible time to finish the scan of a full sweep.
If the data collection platform is in motion during this period, the LiDAR scan may become distorted due to changes in the sensor pose throughout the scan, known as the rolling shutter effect.
To accurately render the LiDAR sweep and model the rolling shutter effect, we interpolate the vehicle pose $\mathbf P_\text{veh} (t)$ for each LiDAR laser at the firing timestamp $t$ and composite it with the sensor extrinsic to obtain the per ray sensor pose, we then generate the LiDAR ray using \cref{eqn:lidar_ray} in main paper for volume rendering.

\begin{figure}[t]
    \centering
     \includegraphics[width=1.0\linewidth]{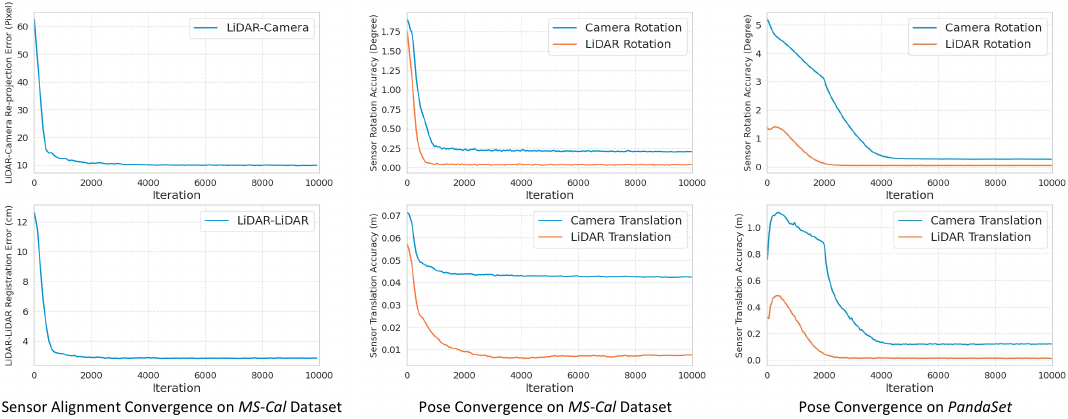}
     \caption{
     \textbf{Convergence of sensor alignment metrics and pose metrics.}
     The calibration typically converges within around $10$K iterations ($\approx 30$ minutes on an A5000 GPU) for both \truckdata and \pandaset datasets.
     \textbf{Left:} Convergence of LiDAR-Camera re-projection error (pixel) on \truckdata checkerboard data and LiDAR-LiDAR registration error (cm) on \truckdata outdoor data.
     \textbf{Middle:} Convergence of sensor pose error on \truckdata dataset.
     \textbf{Right:} Convergence of sensor pose error on \pandaset dataset.
     }
     \label{fig:supp_convergence}
\end{figure}

\paragraph{Surface Alignment Distance:}
To compute the surface alignment distance (\cref{eqn:alignment_loss} in main paper), we first select candidate image pairs that have an overlapping field-of-view.
Then we run SuperPoint~\cite{detone2018superpoint} and LightGlue~\cite{lindenberger2023lightglue} to identify the correspondences between the image pairs.
During training, we randomly select a image for each camera sensor and its candidate pair image to compute the alignment loss in each iteration.
We filter out correspondences with a re-projection error $\lVert \pi(\mathbf p) - \mathbf u \rVert_2 > 50$ for PandaSet~\cite{xiao2021pandaset} and $\lVert \pi(\mathbf p) - \mathbf u \rVert_2 > 20$ for our collected \truckdata dataset.
We also filter out correspondences with a ray termination probability $\sum_{i=1}^N w_i < 0.5$.

\paragraph{Ray Sampling Details:}
During the coarse-to-fine ray sampling phase, we run SuperPoint~\cite{detone2018superpoint} to detect $2048$ keypoints for each camera image and progressively apply Gaussian blur to create the blurred heat maps.
The initial Gaussian blur kernel has a $\sigma=40$ and the final gaussian blur kernel has a $\sigma=5$.

\paragraph{Training Details:}
We employ a multi-stage training schedule.
For the initial $2000$ iterations, the model is trained with uniform ray sampling, and transit to coarse-to-fine ray sampling from iteration $2000$ until the end of training.
Throughout the training process, we utilize the Adam optimizer with a initial learning rate of $0.01$ for the scene model and $0.0001$ for the sensor poses.
The learning rates are exponentially decayed by $0.1$ for the scene model and by $0.01$ for the sensor poses.
During the training, we dynamically adjust the number of sample rays in each iteration to ensure a fixed sample points of $2^{19}$.
The allocation of rays is evenly distributed among sensors to ensure an equal number of sampled rays for each sensor.
Regarding the loss weights in the learning objective, we set $\mathcal L_\text{rgb}=1$, $\mathcal L_\text{int}=0.1$, $\mathcal L_\text{depth}=0.1$, $\mathcal L_\text{align}=0.001$, and $\mathcal L_\text{reg}=0.01$.
We train the model for $30$K iterations in total.
Notably, we observe that calibration typically converges within around 30 minutes on an A5000 GPU.
\cref{fig:supp_convergence} shows the convergences of LiDAR-Camera re-reprojection error, LiDAR-LiDAR registration error, and sensor pose accuracy for the initial $10$k iterations ($\approx 30$ minutes).
However, we opt to extend the training to enhance calibration further.

\begin{table}[t]
    \captionsetup{font=footnotesize}
    \centering
    \begin{tabular}{lccl}
    	\toprule
    	& Run-time $\downarrow$ \\
    	\midrule
    	MLCC & 1 hr \\
        ICP + Edge and Plane + PGO \quad \quad \quad & 3 hr (18 sensor pairs * 10 min)  \\
        \midrule
        SC-NeRF & 1 day \\
        INF & 1 day \\
        Ours & 30 min \\
        \bottomrule
    \end{tabular}
    \caption{Comparison of calibration time on the \truckdata dataset.}
    \label{tab:profiling}
\end{table}

\paragraph{Run-time and Resources:}
Table~\ref{tab:profiling} reports the calibration runtimes compared to other methods.
Our approach is more efficient than the other NeRF-based baselines due to our efficient scene representation and rendering, as well as the surface alignment constraints.
While classical calibration methods are in principle fast, they operate on each sensor pair and the time scales linearly unless parallelized for multi-sensor setups.
Additionally, existing public implementations of classical calibration such as MLCC~\cite{liu2022targetless} or Edge and Plane~\cite{zhou2018automatic} can be slow, complicating direct comparisons.
Besides improved performance, \methodname is more scalable as it does not require expensive infrastructure and operational overhead.
This allows calibration in any location without needing to build calibration sites.

\section{Baseline Implementation Details}
\label{sec:baseline_details}

\subsection{Classical Calibration Baselines}
\paragraph{Point-to-Plane ICP:}
Point-to-Plane ICP~\cite{rusinkiewicz2001efficient} is a commonly used algorithm for LiDAR odometry and LiDAR-LiDAR calibration that is typically more efficient and exhibits better average performance~\cite{pomerleau2013comparing} than point-to-point ICP.
As these algorithms are widely available, we employ the implementation present in Open3D~\cite{zhou2018open3d}.
We run the Point-to-Plane ICP to calibrate each LiDAR pair with sufficient co-visible field-of-views.
The calibration is conducted on the five stationary outdoor scenes that include a variety of poles, walls, and distant objects for calibration.

\paragraph{MLCC:}
MLCC~\cite{liu2022targetless} is a targetless sensor extrinsic calibration method for camera and LiDAR sensors.
We leverage this framework to perform LiDAR-LiDAR calibration.
Specifically, MLCC first uses an adaptive voxelization technique to extract and match LiDAR feature points, subsequently formulating the multi-LiDAR extrinsic calibration problem as a LiDAR Bundle Adjustment (BA) problem.
We employ the implementation from the official repo\footnote{\url{https://github.com/hku-mars/mlcc}} to conduct calibration on our outdoor collects, yielding LiDAR-LiDAR calibration results.
We empirically found the MLCC performs slightly worse than Point-to-Plane ICP~\cite{rusinkiewicz2001efficient}, possibly due to the absence of distinctive structures/features in the outdoor scene data.

\paragraph{Edge and Plane:}
For LiDAR-camera calibration, we use a custom implementation of a commonly-used target-based method~\cite{zhou2018automatic}.
This approach optimizes for both plane and edge correspondences given known dimensions of a checkerboard target.
Improvements over the existing utility available in the MATLAB toolbox involve better target segmentation and edge correspondence matching between identified image edges and extracted point cloud edges.
We exploit more accurate initial CAD estimates for coarse bounding box segmentation and use the known laser scan lines as priors for target edge detection.
This approach ensures we can reject poor matches  in low co-visible regions. To improve robustness, we ensure a variety of target poses are captured and cover the full field-of-view of each camera.
The calibration is performed in an indoor environment to control for lighting conditions and improve target visibility.

\paragraph{Mutual Information:}
Another approach for LiDAR-camera calibration is to leverage the correlation between passive material reflection of visible light and LiDAR reflectivity of near infrared wavelengths.
A popular approach is to maximize the mutual information~\cite{pandey2015automatic} of the grayscale image and the intensity of LiDAR returns. We employ a modified implementation of \cite{ta2023l2e} across 10 stationary outdoor scenes for each co-visible LiDAR-camera pair.
This approach includes image pre-processing steps including a gaussian blur with a standard deviation of 5 pixels and histogram equalization.
We limit the optimization space to within 20 cm and 2 degrees of the initial guess for each translational and rotational DoF.

\paragraph{Pose Graph Optimization:}
To unify LiDAR-camera~\cite{zhou2018automatic}, LiDAR-LiDAR~\cite{chen1992object}, and LiDAR-INS~\cite{furgale2010visual,barfoot2014associating} calibration results, we employ a global pose graph optimization~\cite{zhou2018open3d,choi2015robust} to align the full sensor setup.
As the methods employ different sensing and registration modalities, we unify the pose graph optimization with empirical weights to ensure that traversal of the calibration graph is fully self-consistent.
Pose-graph optimization allows for the averaging across multiple registrations of the same sensors, across sensors, and across registration methods.
As the modalities operate in different domains, however, the process requires careful tuning of the relative information matrices of each edge, which is a tedious and time-consuming process.

\subsection{Neural Rendering Calibration Baselines}

\paragraph{Self-Calibrating NeRF:}
SC-NeRF~\cite{jeong2021self} jointly optimizes the neural field, camera pose, intrinsics, and distortion model. In our experiments, we modified the training algorithm to jointly optimize multiple cameras by randomly sampling a camera at each iteration, sampling rays and computing the loss for the selected camera.
To ensure that the extrinsics can be properly evaluated against the ground truth, we do not optimize the intrinsics and distortion parameters when training.
To match our robot sensor platform setting, we learn a fixed (sensor to vehicle) transformation for each camera from the (given) ground truth vehicle poses at each timestep rather than learning the individual camera to world transformations at each timestep.
In our experiments, we use the NeRF++ \cite{zhang2020nerf++} backbone and the same settings as their Tanks and Temples \cite{knapitsch2017tanks} evaluation with a total of 1.5 million training iterations.
We only replace their scene normalization factor with 100 and 60 for Pandaset~\cite{xiao2021pandaset} and \truckdata respectively to reflect the larger scene sizes.

\paragraph{Implicit Neural Fusion:}
INF~\cite{zhou2023inf} first jointly learns a neural field and pose for a LiDAR sensor, then learns a radiance field and transformation from LiDAR frame to camera frame.
In our experiments, we again modified the training algorithm to learn on multiple LiDARs and multiple cameras by sampling a sensor at random at each training iteration, and sampling rays and computing the loss as usual.
For both LiDARs and cameras, we learn the fixed (sensor to vehicle) transformation from the (given) ground truth vehicle pose at each time step, instead of learning the per-frame sensor to world transformation as was originally done for LiDAR.
We also elected to replace their internal pose representation with 6DoF \cite{zhou2019continuity} as we found that the original pose representation was not well defined for the initial sensor extrinsics.
We also run all experiments with a pinhole camera intrinsics model to match the cameras present in \pandaset and \truckdata datasets.
In our experiments, we base the hyperparameters on the settings for their outdoor scene.
For both datasets, we set the max depth and scene normalization factor for the depth network to 150m.
For Pandaset, we train the density model for 300k iterations and for \truckdata we train for 750k iterations.
For the color model, we set the scene normalization factor to 150m, but set the far range to 50m as we found that this helped with stability.
We also lowered the learning rate of the camera poses to $5e^{-4}$. We train the color model for 1.2 million and 1.6 million iterations for Pandaset and \truckdata respectively, to compensate for the additional cameras present in these datasets.

\begin{figure}[t]
    \centering
     \includegraphics[width=1.0\linewidth]{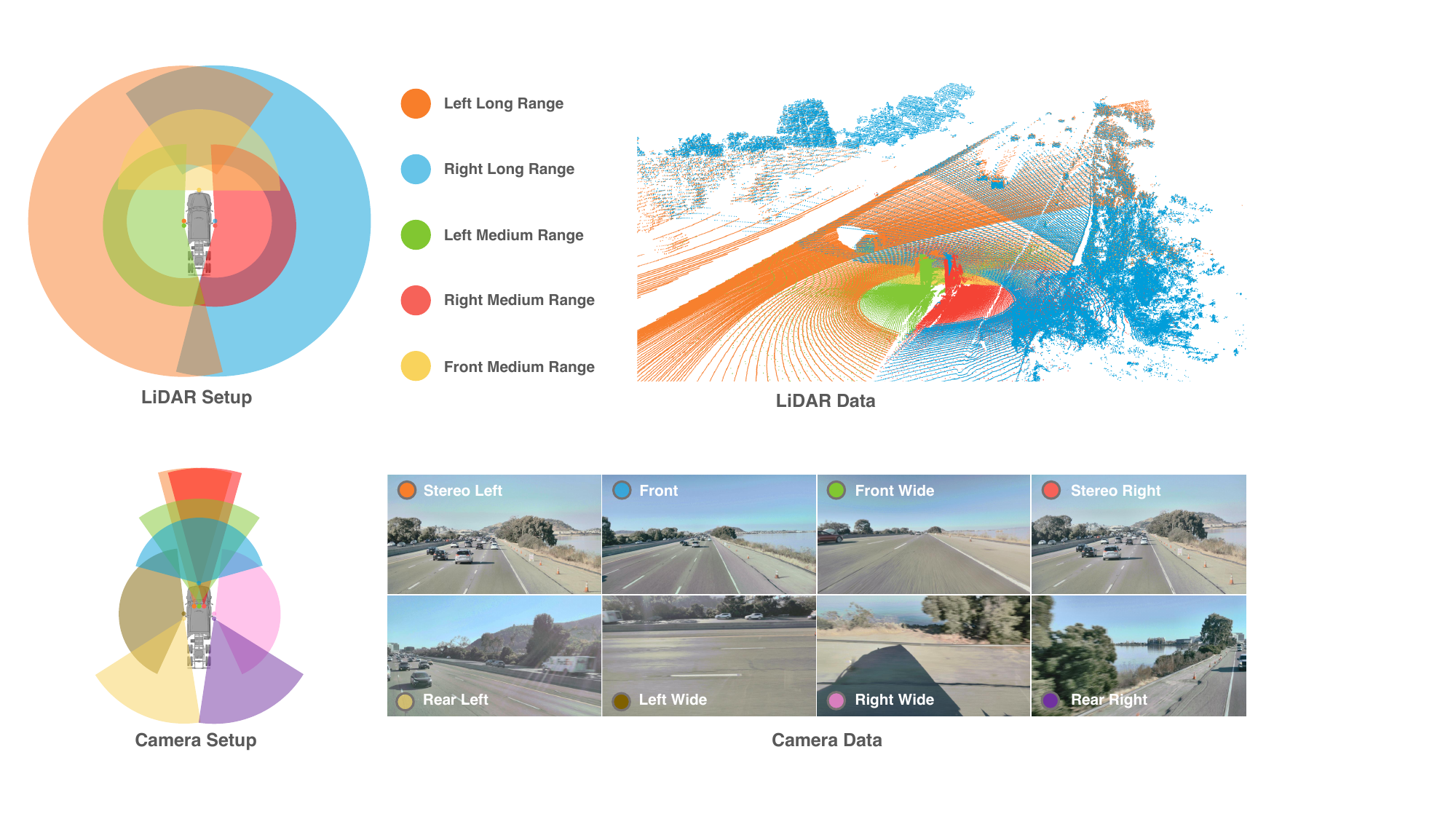}
     \caption{
     \textbf{\truckdata dataset sensor setup and captured data.}
     We show the LiDAR and camera data on the static outdoor collect.
     }
     \label{fig:supp_truck_setup}
\end{figure}

\begin{table*}[t]
    \centering
    \resizebox{1.0\textwidth}{!}{
    \begin{tabular}{lll}
    \toprule
    Camera Type & Camera Name & Paired LiDAR Names \\
    \midrule
    \multirow{2}{*}{Narrow FoV} & Stereo-Left & Long-Range-Left, Long-Range-Right
    \\
    & Stereo-Right & Long-Range-Left, Long-Range-Right
    \\
    \midrule
    \multirow{3}{*}{Medium FoV} & Front & Long-Range-Left, Long-Range-Right
    \\
    & Rear-Left & Long-Range-Left, Medium-Range-Left
    \\
    & Rear-Right & Long-Range-Right, Medium-Range-Right
    \\
    \midrule
    \multirow{3}{*}{Wide FoV} & Front & Long-Range-Left, Long-Range-Right, Medium-Range-Front
    \\
    & Left & Medium-Range-Left
    \\
    & Right & Medium-Range-Right
    \\
    \bottomrule
    \end{tabular}
    }
	\caption{
	\textbf{List of LiDAR-camera pairs for computing re-projection error on \truckdata checkerboard data.}
	}
	\label{tab:supp_lidar_camera_pair}
\end{table*}

\begin{table*}[t]
    \centering
    \resizebox{1.0\textwidth}{!}{
    \begin{tabular}{lcccccccc}
    \toprule
    & \multicolumn{2}{c}{Narrow FoV Cameras} & \multicolumn{3}{c}{Medium FoV Cameras} & \multicolumn{3}{c}{Wide FoV Cameras} \\
    \cmidrule(l){2-3} \cmidrule(l){4-6} \cmidrule(l){7-9}
    & Stereo-Left & Stereo-Right & Front & Rear-Left & Rear-Right & Front & Left & Right
    \\
    \midrule
    Mutual Info~\cite{pandey2012automatic} & 58.75 & 40.10 & 12.59 & 16.87 & 37.46 & 52.55 & 56.50 & 27.89
    \\
    Edge and Plane~\cite{zhou2018automatic} & 11.15 & 9.63 & \textbf{2.28} & \textbf{6.67} & \textbf{12.59} & 11.11 & 14.63 & 11.65
    \\
    Pose Graph Optim~\cite{zhou2018open3d} & 20.46 & 9.68 & 2.89 & 7.77 & 13.75 & 10.14 & \textbf{12.86} & \textbf{11.31}
    \\
    \midrule
    SC-NeRF~\cite{jeong2021self} & 115.81 & 114.98 & 13.83 & 36.82 & 45.09 & 28.66 & 86.82 & 34.17
    \\
    INF~\cite{zhou2023inf} & 54.66 & 24.84 & 3.43 & 15.56 & 63.21 & 57.20 & 120.69 & 61.59
    \\
    \rowcolor{grey} Ours & \textbf{7.78} & \textbf{8.74} & 2.82 & 6.89 & 12.94 & \textbf{9.91} & 15.97 & 12.68
    \\
    \bottomrule
    \end{tabular}
    }
	\caption{
	\textbf{LiDAR-camera re-projection error (in pixel) on \truckdata checkerboard data.}
	We report the breakdown metric for each camera sensor.
	The average metric is in \cref{tab:sota_truck} in main paper.
	}
	\label{tab:supp_lidar_camera_metric}
\end{table*}

\begin{table*}[t]
    \centering
    \resizebox{1.0\textwidth}{!}{
    \begin{tabular}{ll}
    \toprule
    LiDAR Name & Paired LiDAR Name \\
    \midrule
    Long-Range-Left & Long-Range-Right, Medium-Range-Left, Medium-Range-Front
    \\
    Long-Range-Right & Long-Range-Left, Medium-Range-Right, Medium-Range-Front
    \\
    \midrule
    Medium-Range-Front & Long-Range-Left, Long-Range-Right, Medium-Range-Left, Medium-Range-Right
    \\
    Medium-Range-Left & Long-Range-Left, Medium-Range-Front
    \\
    Medium-Range-Right & Long-Range-Right, Medium-Range-Front
    \\
    \bottomrule
    \end{tabular}
    }
	\caption{
	\textbf{List of LiDAR-LiDAR pairs for computing registration error on \truckdata outdoor data.}
	}
	\label{tab:supp_lidar_lidar_pair}
\end{table*}

\begin{table*}[t]
    \centering
    \begin{tabular}{lccccc}
    \toprule
    & \multicolumn{2}{c}{Long Range LiDAR} & \multicolumn{3}{c}{Medium Range LiDAR} \\
    \cmidrule(l){2-3} \cmidrule(l){4-6}
    & Left & Right & Front & Left & Right
    \\
    \midrule
    Point-to-Plane ICP~\cite{chen1992object} & 2.535 & \textbf{2.521} & \textbf{3.056} & 3.042 & \textbf{2.942}
    \\
    MLCC~\cite{liu2022targetless} & 2.842 & 2.829 & 3.215 & 3.318 & 3.194
    \\
    Pose Graph Optim~\cite{zhou2018open3d} & 2.660 & 2.702 & 3.170 & 3.180 & 3.167
    \\
    \midrule
    INF~\cite{zhou2023inf} & 6.559 & 6.800 & 9.626 & 12.526 & 11.158
    \\
    \rowcolor{grey} Ours & \textbf{2.516} & 2.612 & 3.078 & \textbf{3.018} & 3.064
    \\
    \bottomrule
    \end{tabular}
	\caption{
	\textbf{LiDAR-LiDAR registration error (in cm) on \truckdata outdoor data.}
	We report the breakdown metric for each LiDAR sensor.
	The average metric is in \cref{tab:sota_truck} in main paper.
	}
	\label{tab:supp_lidar_lidar_metric}
\end{table*}

\begin{figure}[t]
    \centering
     \includegraphics[width=0.82\linewidth]{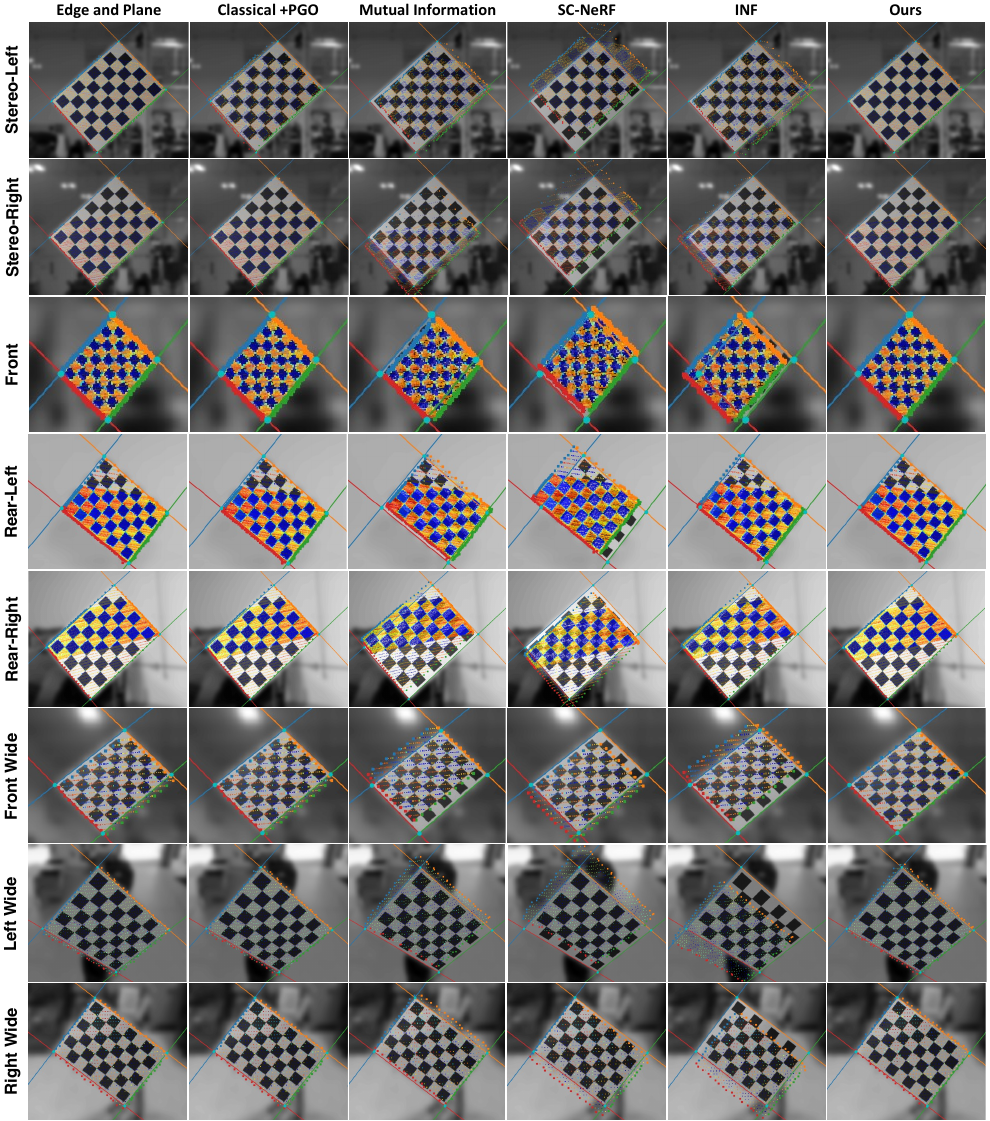}
     \caption{
     \textbf{Visualization of LiDAR-Camera alignment on \truckdata checkerboard data for each camera sensor.}
     We colored the detected checkerboard edge from both LiDAR point cloud and camera images.
     Additionally, the LiDAR points on the checkerboard plane are colored with intensity value.
     }
     \label{fig:supp_checkerboard}
\end{figure}

\begin{figure}[t]
    \centering
     \includegraphics[width=1.0\linewidth]{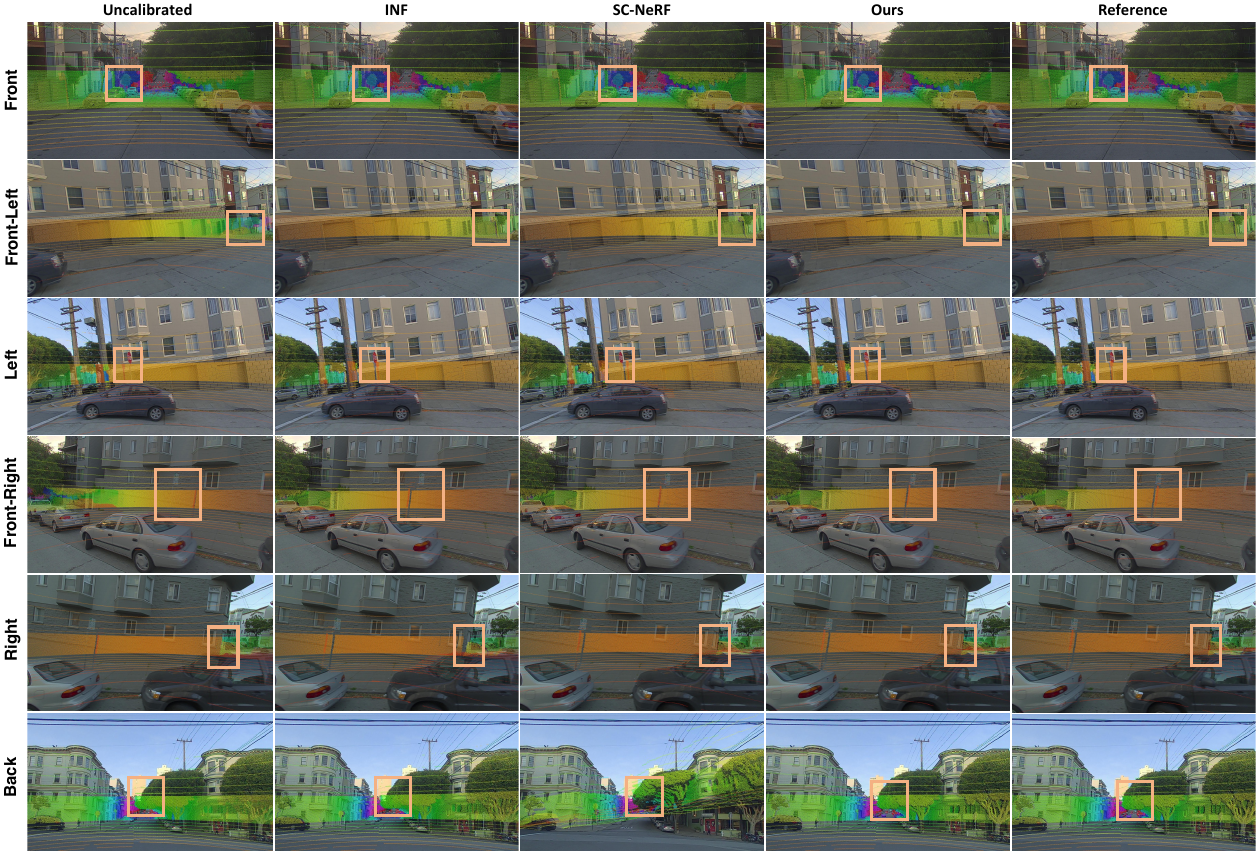}
     \caption{
     \textbf{More qualitative comparison on \pandaset dataset.}
     We show the projection of LiDAR and camera images for each camera sensor. Please zoom-in to see the mis-calibrations.
     }
     \label{fig:supp_panda_qual_1}
\end{figure}

\begin{figure}[t]
    \centering
     \includegraphics[width=1.0\linewidth]{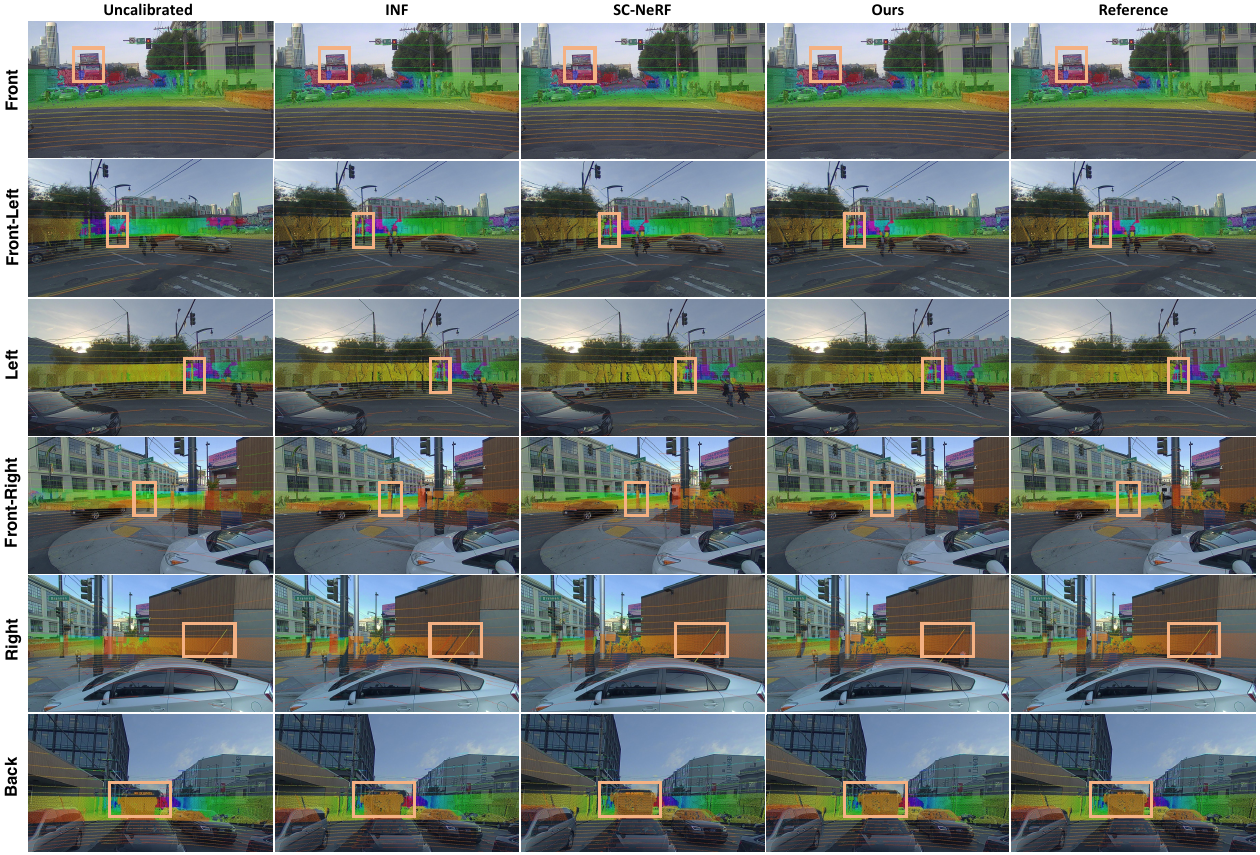}
     \caption{
     \textbf{More qualitative comparison on \pandaset dataset.}
     We show the projection of LiDAR and camera images for each camera sensor. Please zoom-in to see the mis-calibrations.
     }
     \label{fig:supp_panda_qual_2}
\end{figure}

\begin{figure}[t]
    \centering
     \includegraphics[width=1.0\linewidth]{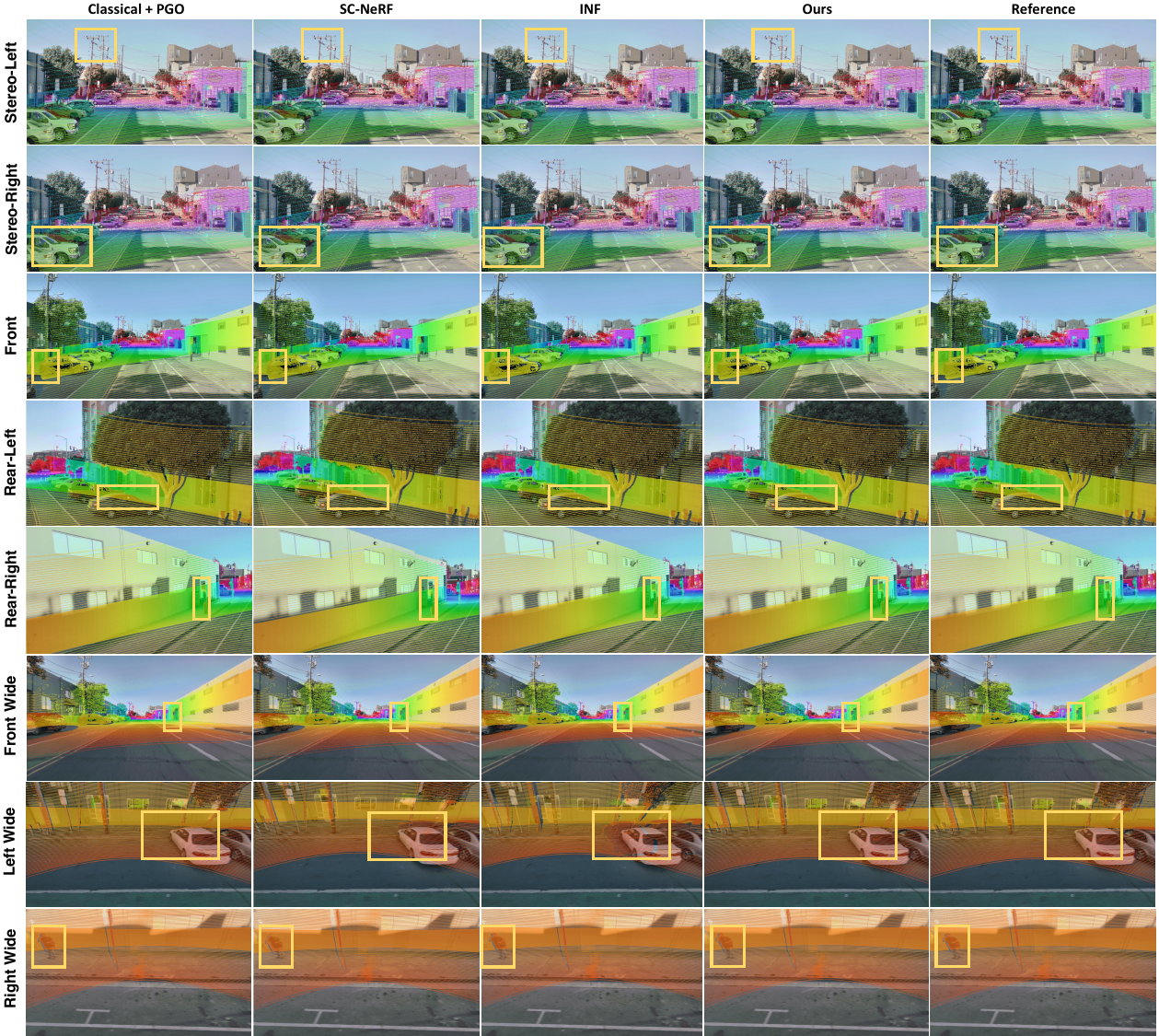}
     \caption{
     \textbf{More qualitative comparison on \truckdata dataset.}
     We show the projection of LiDAR and camera images for each camera sensor. Please zoom-in to see the mis-calibrations.
     }
     \label{fig:supp_truck_qual_1}
\end{figure}

\begin{figure}[t]
    \centering
     \includegraphics[width=1.0\linewidth]{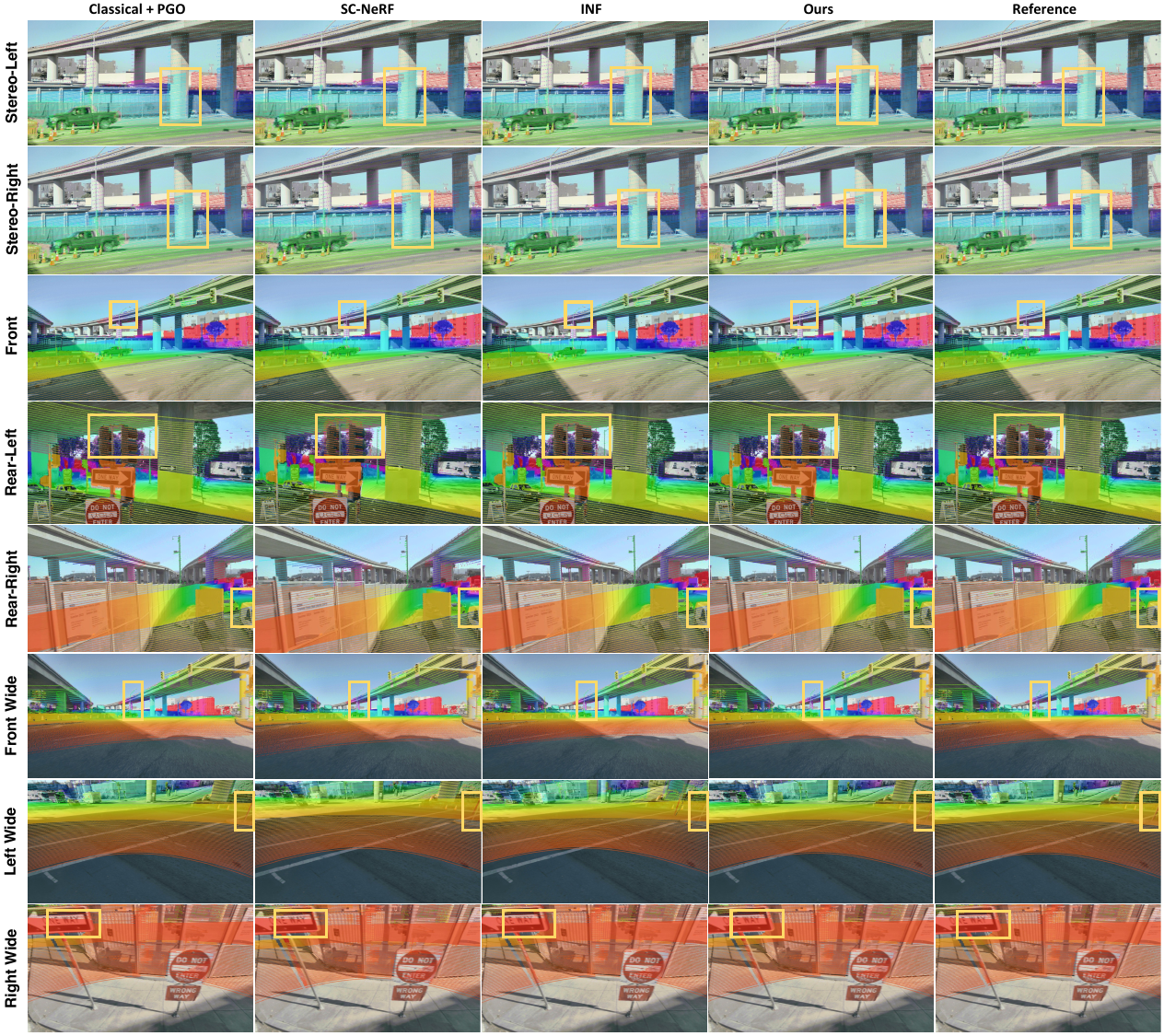}
     \caption{
     \textbf{More qualitative comparison on \truckdata dataset.}
     We show the projection of LiDAR and camera images for each camera sensor. Please zoom-in to see the mis-calibrations.
     }
     \label{fig:supp_truck_qual_2}
\end{figure}

\section{Experiment Details}
\label{sec:exp_details}

\subsection{Multi-Sensor Calibration Dataset}
\begin{figure}[t]
    \centering
     \includegraphics[width=0.9\linewidth]{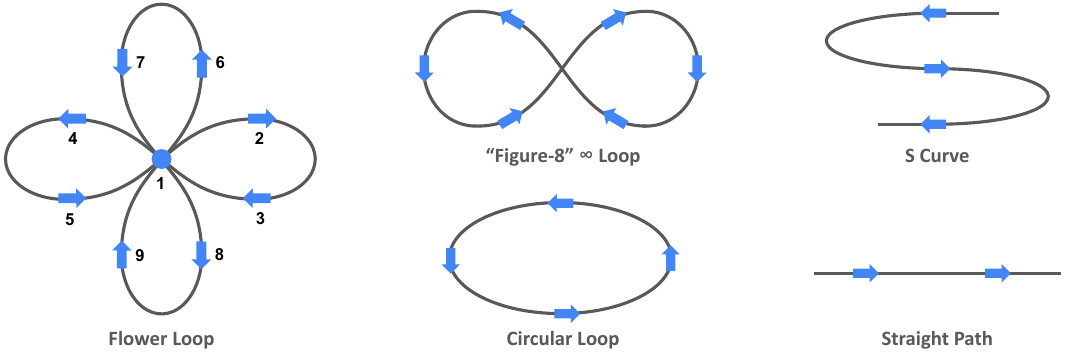}
     \caption{
     \textbf{Illustration of different driving trajectories in \truckdata dataset.}
     Arrows indicate the driving direction, and numbers indicate driving order.
     }
     \label{fig:supp_driving_trajectory}
\end{figure}

We collect a \truckdata dataset to study the calibration performance of our proposed method and baselines.
Additionally, we investigate the impact of driving trajectories on the calibration performance.
The data collection vehicle is a Class 8 truck equipped with five mechanical spinning LiDARs.
Among these, two are long-range LiDARs (with a range of up to $200$m) positioned on the left and right sides of the truck.
The remaining three are medium-range LiDARs (with a range of up to $50$m) mounted at the front, left, and right sides of the truck to provide near-range sensing.
The LiDAR setup ensures comprehensive $360^\circ$ coverage.
In addition to LiDARs, the data collection vehicle is equipped with eight cameras.
These include three wide-angle cameras oriented towards the front, left, and right.
Furthermore, three medium-angle cameras are placed to capture views from the front, rear left, and rear right.
Lastly, two long-range stereo cameras are positioned at the front to provide far-distant observations.
Please refer to \cref{fig:supp_truck_setup} for a visual representation of the sensor setup and the data captured.
The dataset includes indoor data and outdoor data.
For the indoor data, the data collection vehicle remains stationary, while a checkerboard is positioned at various locations and orientations for camera-LiDAR pair calibration.
One collect is utilized for optimizing classical calibration methods, while another is held out for evaluating LiDAR-camera sensor alignment metrics.
For the outdoor parking lot data, we collected both stationary and dynamic trajectories, including eight stationary scenes and four "figure-8" $\infty$ loops.
Two $\infty$ loops were selected for training neural-rendering methods, and five stationary scenes were used for classical LiDAR alignment calibration.
The remaining $\infty$ loops and stationary scenes were reserved for evaluating the calibrations.
To delve into the influence of driving trajectories on calibration performance, we also collect six additional outdoor dynamic trajectories.
These trajectories include two flower loops, two circular loops, one S curve, and one straight path.
Please refer to \cref{fig:supp_driving_trajectory} for an illustration of different trajectories.

\subsection{Urban Driving Dataset}
To evaluate our method on urban driving dataset, we choose public available real-world \pandaset~\cite{xiao2021pandaset}, which contains $103$ urban driving scenes captured in San Francisco.
Each scene spans $8$ seconds, equivalent to $80$ frames sampled at $10$Hz.
The data collection platform consists of a $360^\circ$ mechanical spinning LiDAR as well as a forward-facing LiDAR, along with six cameras.
These cameras are facing front, front-left, left, back, front-right, and right.
We calibrate all the sensors, including the two LiDARs and six cameras.
Please see
\cref{fig:teaser} in main paper
for the sensor setup.
To quantitatively evaluate our approach against baseline methods that are computationally intensive to train, we selected scenes that have few dynamic actors as the calibration logs.
Our selected logs also have different driving trajectories (\eg incline, turning) and feature rich geometric elements in the scene (\eg parked vehicles).
We selected four logs \texttt{028, 039, 040, 053} for calibration training.
We chose two scenes for reconstruction evaluation: \texttt{034, 056}.
This necessitated the training of eight reconstruction and rendering models for each baseline.

\subsection{Reference Calibration for Pose Accuracy Metrics}
To evaluate the pose accuracy metrics, we report the average rotation and translation error between the reference and the estimated calibrations.
We now describe how we obtain the reference calibration for \pandaset and \truckdata datasets.

\paragraph{\pandaset Dataset:}
For \pandaset~\cite{xiao2021pandaset}, we use their provided calibration file\footnote{\url{https://github.com/scaleapi/pandaset-devkit/blob/master/docs/static_extrinsic_calibration.yaml}} as the reference for ground truth.
It is noted that this file exclusively contains relative poses between different sensors, but does not provide a reference pose of the sensors to the vehicle.
To establish the pose between the sensors and the vehicle frame of reference, we run Iterative Closest Point (ICP) algorithm between the raw LiDAR (in sensor coordinates) and the pose-processed LiDAR (in vehicle coordinates) on the static log \texttt{004}.
This process enabled us to determine the $\mathbb{SE}(3)$ transform between the $360^\circ$ mechanical spinning LiDAR and the vehicle frame.
In log \texttt{004}, the data collection vehicle remains stationary, eliminating rolling shutter effects.
The computed rotation from the $360^\circ$ mechanical spinning LiDAR to the vehicle frame (FLU convention) is represented in quaternion as:
\texttt{\{w:-6.9577e-01, x:5.8054e-03, y:5.2777e-03, z:-7.1823e-01\}}.
The translation is given by:
\texttt{\{x:7.8202e-01, y:1.1396e-04, z:1.8596e+00\}} in meters.

\paragraph{\truckdata Dataset:}
For our collected \truckdata dataset, we utilized both classical calibration and brute-force blackbox optimization to establish the ground-truth reference.
We first calibrate each LiDAR-LiDAR pair using Point-to-Plane ICP~\cite{chen1992object} on the outdoor stationary collects, and we calibrate each LiDAR-camera pair using Edge and Plane~\cite{zhou2018automatic} correspondences on the indoor checkerboard colects, and we calibrate the long-range LiDAR to Inertial Navigation System (INS) based on LiDAR odometry~\cite{furgale2010visual,barfoot2014associating}.
Subsequently, we run pose graph optimization to derive the optimal global alignment for full sensor calibration.
Finally, we run blackbox optimization~\cite{powell1964efficient} to search the reference calibration that minimizes both LiDAR-camera re-projection error (evaluated on the indoor checkerboard data) and LiDAR-LiDAR registration error (evaluated on the outdoor static data) on the evaluation collects.
The search space for optimization was identified by analyzing the range of pose discrepancies between the different evaluated calibration methods.

\begin{table}[t]
     \centering
     \begin{tabular}{lcccc}
     \toprule
     \multirow{2}{*}{Method} & \multicolumn{2}{c}{Stereo-Left Camera} & \multicolumn{2}{c}{Stereo-Right Camera} \\
     \cmidrule(l){2-3} \cmidrule(l){4-5}
     & Rotation$\downarrow$ & Translation$\downarrow$ & Rotation$\downarrow$ & Translation$\downarrow$ \\
     \midrule
     CalibNet~\cite{iyer2018calibnet} & $5.83 \pm 2.93 ^\circ$ & $14.36 \pm 6.37$ cm & $5.77 \pm 2.93 ^\circ$ & $14.22 \pm 6.37$ cm
     \\
     LCCNet~\cite{lv2021lccnet} & $0.17 \pm 0.45 ^\circ$ & ~~$\mathbf{1.29 \pm 1.94}$ cm  & $1.52 \pm 0.71 ^\circ$ & $52.49 \pm 0.51$ cm
     \\
     Ours & $\mathbf{0.12} \pm \mathbf{0.07} ^\circ$ & ~~$2.16 \pm 0.49$ cm  &  $\mathbf{0.12} \pm \mathbf{0.05} ^\circ$ & ~~$\mathbf{1.96} \pm \mathbf{0.87}$ cm
     \\
     \bottomrule
     \end{tabular}
 	\caption{
 	\textbf{Comparison of calibration accuracy to learning-based methods on \textit{KITTI-odometry} dataset}.
 	$10^\circ$ rotation error and $20$ cm translation error are added on each axis.
 	CalibNet~\cite{iyer2018calibnet} and LCCNet~\cite{lv2021lccnet} are both trained on stereo left camera, we also report metrics on stereo right camera.
 	}
 	\label{tab:kitti_comparison}
\end{table}

\begin{table}[t]
     \centering
     \resizebox{1.0\textwidth}{!}{
     \begin{tabular}{llcccc}
     \toprule
     \multirow{2}{*}{Perturbation} & \multirow{2}{*}{~~Method} & \multicolumn{2}{c}{Front-Right Camera} & \multicolumn{2}{c}{LiDAR} \\
     \cmidrule(l){3-4} \cmidrule(l){5-6}
     & & Rotation$\downarrow$ & Translation$\downarrow$ & Rotation$\downarrow$ & Translation$\downarrow$ \\
     \toprule
     \multirow{2}{*}{Rotation $2^\circ$} & ~~MOISST~~ & $0.09 \pm 0.01 ^\circ$ & $1.6 \pm 0.3$ cm & $0.39 \pm 0.09 ^\circ$ & $9.2 \pm 1.9$ cm
     \\
     & ~~Ours~~ & $\mathbf{0.06 \pm 0.01} ^\circ$ & $\mathbf{0.9 \pm 0.2}$ cm & $\mathbf{0.26 \pm 0.05} ^\circ$ & $\mathbf{1.5 \pm 0.4}$ cm
     \\
     \cmidrule(l){2-6}
     \multirow{2}{*}{Rotation $5^\circ$} & ~~MOISST~~ & $0.07 \pm 0.05 ^\circ$ & $1.7 \pm 0.5$ cm & $0.40 \pm 0.14 ^\circ$ & $9.9 \pm 2.3$ cm
     \\
     & ~~Ours~~ & $\mathbf{0.07 \pm 0.01} ^\circ$ & $\mathbf{0.9 \pm 0.2}$ cm & $\mathbf{0.31 \pm 0.08} ^\circ$ & $\mathbf{1.8 \pm 0.5}$ cm
     \\
     \cmidrule(l){2-6}
     \multirow{2}{*}{Rotation $10^\circ$} & ~~MOISST~~ & $15.81 \pm 0.02 ^\circ$ & $127.2 \pm 0.6$ cm & $17.07 \pm 0.13 ^\circ$ & $104.4 \pm 1.7$ cm
     \\
     & ~~Ours~~ & $\mathbf{0.11 \pm 0.04} ^\circ$ & $\mathbf{1.6 \pm 0.6}$ cm & $\mathbf{0.37 \pm 0.20} ^\circ$ & $\mathbf{1.8 \pm 0.5}$ cm
     \\
     \midrule
     \multirow{2}{*}{Transl $20$ cm} & ~~MOISST~~ & $0.09 \pm 0.02 ^\circ$ & $1.8 \pm 0.4$ cm & $0.46 \pm 0.1 ^\circ$ & $8.7 \pm 1.5$ cm
     \\
     & ~~Ours~~ & $\mathbf{0.06 \pm 0.01} ^\circ$ & $\mathbf{1.2 \pm 0.2}$ cm & $\mathbf{0.19 \pm 0.02} ^\circ$ & $\mathbf{2.1 \pm 0.5}$ cm
     \\
     \cmidrule(l){2-6}
     \multirow{2}{*}{Transl $50$ cm} & ~~MOISST~~ & $0.09 \pm 0.02 ^\circ$ & $1.7 \pm 0.5$ cm & $0.5 \pm 0.06 ^\circ$ & $7.8 \pm 1.2$ cm
     \\
     & ~~Ours~~ & $\mathbf{0.06 \pm 0.01} ^\circ$ & $\mathbf{1.1 \pm 0.3}$ cm & $\mathbf{0.20 \pm 0.01} ^\circ$ & $\mathbf{2.3 \pm 0.9}$ cm
     \\
     \cmidrule(l){2-6}
     \multirow{2}{*}{Transl $100$ cm} & ~~MOISST~~ & $0.09 \pm 0.02 ^\circ$ & $1.7 \pm 0.3$ cm & $0.43 \pm 0.08 ^\circ$ & $8.8 \pm 2.4$ cm
     \\
     & ~~Ours~~ & $\mathbf{0.06 \pm 0.01} ^\circ$ & $\mathbf{1.2 \pm 0.2}$ cm & $\mathbf{0.21 \pm 0.02} ^\circ$ & $\mathbf{2.5 \pm 0.7}$ cm
     \\
     \bottomrule
     \end{tabular}
     }
 	\caption{
 	\textbf{Comparison of calibration accuracy to MOISST~\cite{herau2023moisst} on \textit{KITTI-360} dataset with different calibration initialization}.
 	Our method can recover from large rotational error ($10^\circ$) while MOISST failed to get a satisfactory calibration.
 	}
 	\label{tab:supp_perturbation}
\end{table}

\begin{table}[t]
     \centering
     \resizebox{1.0\textwidth}{!}{
     \begin{tabular}{lccccccc}
     \toprule
     \multirow{2}{*}{Driving Trajectory} & \multicolumn{2}{c}{Camera Pose} & \multicolumn{2}{c}{LiDAR Pose} & \multicolumn{3}{c}{Rendering Quality}\\
     \cmidrule(l){2-3} \cmidrule(l){4-5} \cmidrule(l){6-8}
     & Rotation$\downarrow$ & Translation$\downarrow$ & Rotation$\downarrow$ & Translation$\downarrow$ & PSNR$\uparrow$ & SSIM$\uparrow$ & Depth$\downarrow$ \\
     \midrule
     Straight path & 0.374 & 0.046 & 0.075 & 0.013 & 29.59 & 0.845 & 0.137
     \\
     Circular loop & 0.271 & 0.098 & 0.065 & 0.030 & 31.75 & 0.898 & 0.040
     \\
     S curve & 0.210 & 0.050 & 0.054 & \textbf{0.008} & 31.83 & 0.899 & 0.037
     \\
     $\infty$ loop & 0.186 & \textbf{0.033} & \textbf{0.036} & \textbf{0.008} & 31.96 & 0.903 & \textbf{0.035}
     \\
     Flower loop & \textbf{0.178} & 0.041 & 0.039	 & 0.009 & \textbf{31.97} & \textbf{0.904} & \textbf{0.035}
     \\
     \bottomrule
     \end{tabular}
     }
 	\caption{\textbf{Analysis of various driving trajectories} on \truckdata dataset.
 	Rotational errors are measured in degrees, while translation errors are measured in meters.
 	}
 	\label{tab:supp_driving_trajectory}
\end{table}

\begin{table}[t]
     \centering
     \resizebox{1.0\textwidth}{!}{
     \begin{tabular}{lccccccc}
     \toprule
     \multirow{2}{*}{Correspondance Loss} & \multicolumn{2}{c}{Camera Pose} & \multicolumn{2}{c}{LiDAR Pose} & \multicolumn{3}{c}{Rendering Quality}\\
     \cmidrule(l){2-3} \cmidrule(l){4-5} \cmidrule(l){6-8}
     & Rotation$\downarrow$ & Translation$\downarrow$ & Rotation$\downarrow$ & Translation$\downarrow$ & PSNR$\uparrow$ & SSIM$\uparrow$ & LPIPS$\downarrow$ \\
     \midrule
     No & 0.856 & 0.614 & \textbf{0.047} & \textbf{0.015} & 23.89 & 0.681 & 0.500
     \\
     Projected Ray Dist & 0.550 & 0.622 & \textbf{0.047} & \textbf{0.015} & 24.02 & 0.689 & 0.492
     \\
     Surface Alignment Dist & \textbf{0.267} & \textbf{0.122} & 0.048 & \textbf{0.015} & \textbf{25.14} & \textbf{0.727} & \textbf{0.450}
     \\
     \bottomrule
     \end{tabular}
     }
 	\caption{
 	\textbf{Comparison of Surface Alignment Distance and Projected Ray Distance} on \pandaset dataset.
 	Rotational errors are measured in degrees, while translation errors are measured in meters.
 	}
 	\label{tab:supp_projected_ray_distance}
\end{table}

\begin{table}[t]
     \centering
     \begin{tabular}{lcccc}
     \toprule
     \multirow{2}{*}{Sensors} & \multicolumn{2}{c}{Camera Pose} & \multicolumn{2}{c}{LiDAR Pose} \\
     \cmidrule(l){2-3} \cmidrule(l){4-5}
     & Rotation$\downarrow$ & Translation$\downarrow$ & Rotation$\downarrow$ & Translation$\downarrow$ \\
     \midrule
     Full & \textbf{0.267} & \textbf{0.122} & \textbf{0.048} & \textbf{0.015} \\
     Camera-only & 0.308 & 0.133 & - & -
     \\
     LiDAR-only & - & - & 0.050 & 0.017
     \\
     \bottomrule
     \end{tabular}
 	\caption{
 	\textbf{Camera-only and LiDAR-only calibration results} on \pandaset.
 	Rotational errors are measured in degrees, while translation errors are measured in meters.
 	}
 	\label{tab:supp_single_modality}
\end{table}

\begin{figure}[t]
    \centering
     \includegraphics[width=1.0\linewidth]{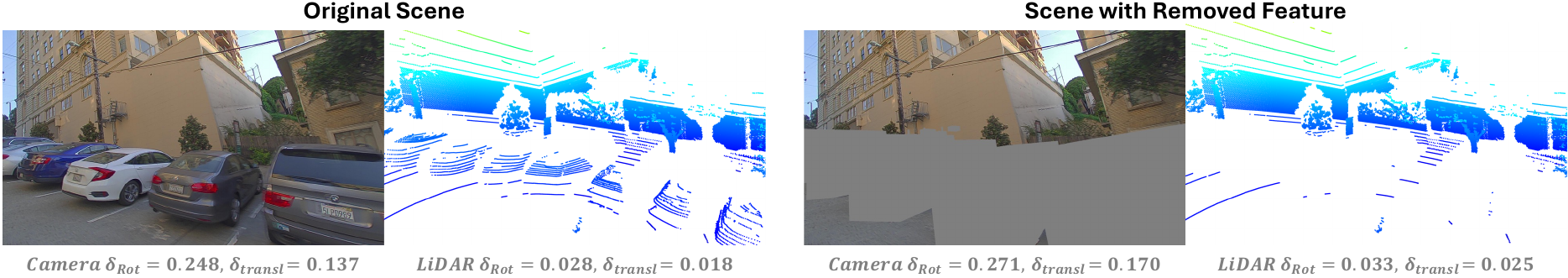}
     \caption{
     We remove LiDAR points and camera pixels of objects from existing scenes to simulate the change of scene featureness.
     \textbf{Left:} Camera image and LiDAR point cloud of original scene.
     \textbf{Right:} Scene after removing actors, with corresponding camera image and LiDAR point cloud.
     We show the Camera and LiDAR pose error \wrt the reference below the figures.
     }
     \label{fig:scene_feature_ness}
\end{figure}

\begin{table}[t]
    \centering
	\begin{tabular}{lccccc}
    \toprule
    {Rendering Model} & {Calibration} & {PSNR$\uparrow$ } & {SSIM$\uparrow$ } & {LPIPS$\downarrow$ } & {Depth$\downarrow$ } \\
    \midrule
    \multirow{2}{*}{Rasterization~\cite{kerbl20233d}}
    & Original & 20.91 & 0.661 & 0.526 & -
    \\
    & \methodname & \textbf{21.27} & \textbf{0.666} & \textbf{0.523} & -
    \\
    \midrule
    \multirow{2}{*}{Raytracing~\cite{yang2023unisim}}
    & Original & 24.54 & 0.696 & 0.474 & \textbf{0.049}
    \\
    & \methodname & \textbf{25.23} & \textbf{0.730} & \textbf{0.449} & \textbf{0.049}
    \\
    \bottomrule
    \end{tabular}
	\caption{
	\textbf{Scene reconstruction and rendering} with dataset original and \methodname-refined calibration for rasterization-based 3D-GS~\cite{kerbl20233d} and raytracing-based UniSim~\cite{yang2023unisim}.
	}
	\label{tab:reconstruction}
\end{table}

\subsection{Evaluation Metric Details}
For the \textbf{LiDAR-camera alignment metric}, we report the average re-projection error measured in pixels between the corners of the checkerboard planes derived from the LiDAR points and those in the images.
This assessment is conducted in a $1080 \times 1920$ resolution image.
For the \textbf{LiDAR-LiDAR alignment metric}, we compute the average Point-to-Plane distance (cm) for all inlier correspondences for each LiDAR pair on the stationary evaluation scenes.
To identify inlier correspondences, we set a maximum correspondence distance of $30$cm.
Regarding the \textbf{pose accuracy metrics}, we report the average rotation error (in degree) and translation error (in meter) between the reference calibration and the estimated calibration.
Since some of the baseline methods we compare against do not perform calibration with respect to a reference point on the vehicle, we designate a root sensor and align its calibrated pose with its reference before computing the metric.
Specifically, for \pandaset, the root sensor is set as the $360^\circ$ mechanical spinning LiDAR, while for the \truckdata dataset, it is the long-range mechanical spinning LiDAR.
For the \textbf{rendering metrics}, we train a neural rendering model~\cite{yang2023unisim} for each rendering scene, considering the calibration result from each calibration scene.
This entails training a total of $N_\text{calib} \times N_\text{render}$ neural rendering models for each baseline.
Each model is trained on every other frame and evaluated on the remaining frames.
The reported rendering metrics represent the averages across the $N_\text{calib} \times N_\text{render}$ models for each baseline.
Note that to ensure fair comparison, the neural rendering method is fixed across all methods, and only the input calibration from each evaluated method changes -
We optimizing the neural rendering model given the evaluated calibration result.

\section{Additional Experiments and Analysis}
\label{sec:additional_results}
In this section, we provide additional quantitative and qualitative results, additional comparison to learning-based (CalibNet~\cite{iyer2018calibnet}, LCCNet~\cite{lv2021lccnet}) and NeRF-based (MOISST~\cite{herau2023moisst}) methods, analysis on the calibration initialization and driving trajectory, the feature-ness of the scenes, and additional ablation study.
We also show that \methodname improved calibration enables more realistic reconstruction and simulation of driving scenes.

\subsection{Additional Sensor Alignment Results}
\cref{tab:supp_lidar_camera_pair} shows all the LiDAR-camera pairs used to compute re-projection error, we report the LiDAR-camera re-projection error for each camera in \cref{tab:supp_lidar_camera_metric}.
Additionally, \cref{tab:supp_lidar_lidar_pair} shows all the LiDAR-LiDAR pairs used to compute registration error, and the corresponding LiDAR-LiDAR registration error for each LiDAR is detailed in \cref{tab:supp_lidar_lidar_metric}.
Please refer to \cref{fig:supp_checkerboard} for a visual comparison of the LiDAR-camera alignment on the \truckdata checkerboard data for each camera sensor.
It can be seen from the figure that our method consistently achieves better sensor alignment compared to baseline methods across all sensors.

\subsection{Additional Qualitative Results}
For more qualitative comparisons of projections of LiDAR points and camera images, please refer to \cref{fig:supp_panda_qual_1} and \cref{fig:supp_panda_qual_2} for examples from the \pandaset dataset.
Additionally, we show qualitative results from our collected data on urban driving scenes in \cref{fig:supp_truck_qual_1} and \cref{fig:supp_truck_qual_2}.

\subsection{Additional Comparison with Learning-based Methods}
Learning-based approaches~\cite{lv2021lccnet,schneider2017regnet,iyer2018calibnet,wu2021netcalib} formulate extrinsic prediction from camera and LiDAR observations as a supervised learning task.
They are effective on the trained sensor configurations/scenes similar to those scene in training and are fast to run.
We compare to LCCNet~\cite{lv2021lccnet}\footnote{\url{https://github.com/IIPCVLAB/LCCNet}} and CalibNet~\cite{iyer2018calibnet}\footnote{\url{https://github.com/gitouni/CalibNet_pytorch}} using the provided pre-trained model.
We follow the same setting as in~\cite{lv2021lccnet,iyer2018calibnet} to use the odometry branch of the KITTI~\cite{kitti} dataset.
\cref{tab:kitti_comparison} shows the results on sequence 00.
LCCNet and CalibNet pre-trained models are trained on stereo left camera, we report the calibration accuracy results on both stereo-left camera and stereo-right camera by intializeing the calibration with rotation error perturbations of $10^\circ$ and translation errors of $20$ cm on each axis.
We use $10$ different seeds and compute the error statistics over these $10$ runs.
LCCNet's performance is good on trained stereo left camera, but degrades on unseen stereo right camera, indicating that learning-based methods do require re-training when the sensor configuration changes and also requires access to the GT calibration for training.

\subsection{Analysis on Calibration Initialization and Comparison to NeRF-based Method MOISST \cite{herau2023moisst}}
We also study the calibration initialization and compare to MOISST~\cite{herau2023moisst}.
Specifically, we follow the same setting as in MOISST~\cite{herau2023moisst} and report results on KITTI-360~\cite{liao2022kitti} NVS training sequence 1.
We consider the front-left (stereo-left) camera as reference sensor and apply up to $\pm 100$ cm translation and $\pm 10^\circ$ rotation offsets on all axes to simulate spatial calibration errors, respectively.
For each perturbation level, we use $10$ different seeds and compute the error statistics over these runs.
\cref{tab:supp_perturbation} shows the calibration results and comparison to MOISST~\cite{herau2023moisst} with different translation perturbation and rotation perturbation initialization.
Our method can recover accurate calibration from large rotational and translation errors compared to MOISST~\cite{herau2023moisst} due to our additional calibration-inspired enhancements, such as surface alignment constraints.

\subsection{Analysis on Driving Trajectory}
To study the calibration performance on different driving patterns, we run our method on the straight path, circular loop, S curve, $\infty$ loop, and flower loop on our collected \truckdata dataset.
\cref{tab:supp_driving_trajectory} shows the results for each driving trajectory.
It can be seen from the tables that straight path and circular loop exhibit inferior performance compared to other trajectories, possibly due to under-constrained observations and incomplete sensor overlap.
This implies that running $\infty$ or flower loops is more favorable for multi-sensor calibration.

\subsection{Performance on Feature-less Scenes}
Our method assumes that there is interesting scene geometry with which to reconstruct, and may have challenges on empty scenes with little to no geometry features.
We analyze our method's performance when reducing features in the scene by removing annotated actor observations from an existing scene (Figure~\ref{fig:scene_feature_ness}).
Specifically, we leverage annotated bounding boxes to identify actors within the scene (\eg vehicles, motorcycles, pedestrians, and construction items), and subsequently remove corresponding camera pixels and LiDAR points.
Figure~\ref{fig:scene_feature_ness} shows a comparison of UniCal's performance on the original scene and the scene with removed features, utilizing PandaSet \texttt{Log-040}.
The performance drop is small even upon the removal of all annotated objects within the scene.

\subsection{Additional Ablation Study}
We further study the effectiveness of the surface alignment loss
(\cref{eqn:alignment_loss} in the main paper)
as compared to the projected ray distance proposed in SC-NeRF~\cite{jeong2021self}.
The projected ray distance measures the distance between the corresponding rays from pairs of camera images but falls short in ensuring that the 3D structure inferred from these correspondences aligns accurately with the underlying scene representation.
\cref{tab:supp_projected_ray_distance} presents a comparison of surface alignment distance and projected ray distance.
The table reveals that optimizing the projected ray distance alone faces challenges in recovering accurate camera rotation and translation.
Additionally, we show the camera-only and LiDAR-only calibration results in \cref{tab:supp_single_modality}.
The results demonstrate that leveraging multiple sensor modalities leads to better performance.

\subsection{\methodname Improves Scene Reconstruction}
We find that with \methodname, we can further refine the calibration from the existing reference provided by \pandaset to achieve better scene reconstruction and rendering.
We jointly learn the calibration using the calibration logs on \pandaset to obtain the refined sensor calibration, and compare this refined calibration with the original calibration on evaluation logs for novel view synthesis.
Table~\ref{tab:reconstruction} shows that for both raytracing-based UniSim~\cite{yang2023unisim} model and rasterization-based 3D Gaussian Splatting~\cite{kerbl20233d} model, \methodname's refined calibration consistently leads to better scene reconstruction and novel viewpoint rendering.

\section{Limitations and Future Works}
\label{sec:limitation}
Our method focuses on calibrating the sensor extrinsics offline and currently assumes that the intrinsics and trajectory are provided.
We also focus on calibrating using static scenes and do not explicitly model changes in lighting or motion.
We note that we focus on calibration of LiDAR and camera sensors and exclude calibration of other sensors, such as the IMU sensor.
As noted in Table~\ref{tab:supp_driving_trajectory}, our method also has performance variation depending on the trajectory driven.
\methodname also assumes that there is interesting scene geometry with which to reconstruct, and may have challenges on empty scenes with little to no geometry features.
Future work will involve extending the method to reduce these assumptions for further robustness and scalability.

\paragraph{Potential Negative Social Impact:}
Our methods are valuable for self-driving sensor calibration.
We acknowledge that there might be privacy concerns arising from data collection used for running \methodname, which can be mitigated through data anonymization techniques.
While \methodname significantly reduces costs and operational overhead for calibrating large SDV fleets, we recognize that there may be situations where the calibration results deviate from the reality.
Holistic and thorough evaluation of autonomy safety before deploying is critical.

\end{document}